%% file: main.tex
\pdfoutput=1
\documentclass{article}

\usepackage[main,final]{neurips_2026}

\usepackage[utf8]{inputenc}
\usepackage[T1]{fontenc}
\usepackage{hyperref}
\usepackage{url}
\usepackage{amsfonts}
\usepackage{nicefrac}
\usepackage{microtype}
\usepackage{graphicx}
\usepackage{amsmath}
\usepackage{enumitem}
\usepackage{amssymb}
\usepackage{caption}
\usepackage{booktabs}
\usepackage{multirow}
\usepackage{makecell}
\usepackage[table]{xcolor}
\usepackage{adjustbox}
\usepackage{amsthm}
\usepackage{listings}
\newtheorem{proposition}{Proposition}

\usepackage{colortbl}
\usepackage{caption}
\definecolor{ourscolor}{gray}{0.92}

\hypersetup{pdftitle={MissPath-FM: Flow Matching with Structured Priors for Partially Observed Time Series}, pdfauthor={Genpei Zhang}}

\title{MissPath-FM: Flow Matching with Structured Priors for Partially Observed Time Series}

\author{%
  Genpei Zhang \\
  Department of Computer Sciences\\
  University of Wisconsin--Madison\\
  \texttt{genpei.zhang@wisc.edu} \\
}

\begin{document}
\maketitle

\input{sec/0_abstract}
\input{sec/1_intro}
\input{sec/2_related}
\input{sec/3_method_nips}
\input{sec/4_experiment_nips}
\input{sec/5_results_nips}
\input{sec/6_conclusion}

\begin{ack}
The author thanks the anonymous reviewers and the area chair for their constructive feedback, which substantially improved the clarity of this paper.
The author received no specific funding for this work and declares no competing interests.
\end{ack}

\bibliographystyle{plain} %
\bibliography{references}

\newpage
\appendix
\input{sec/7_appendix}

\end{document}

%% file: sec/0_abstract.tex
\begin{abstract}
Flow matching has emerged as a powerful alternative to diffusion models for generative modeling, but standard conditional flow matching is fundamentally misaligned with partially observed time series: starting from an uninformed Gaussian source forces the model to transport even already-observed entries from noise to the target, wasting capacity on trivially known regions.
We propose \textbf{MissPath-FM}, a partial-observation-aware flow matching framework that redesigns the source prior and probability path for incomplete time series.
MissPath-FM employs \emph{structured source priors} (interpolation priors and Gaussian process posterior priors) to initialize the flow close to the observed signal, paired with a \emph{missingness-aware smooth path} that allocates transport asymmetrically between observed and missing regions.
We evaluate systematically across four datasets, four missingness mechanisms (MCAR, block, MNAR, irregular), two missing ratios, and three sequence lengths ($T \in \{24, 48, 96\}$), totaling 48 settings against eight baselines.
MissPath-FM wins 29 of 32 settings at $T{=}48$ and 45 of 48 overall (93.8\%).
Our ablations reveal a regime-dependent design hierarchy in which no single component suffices: under random missingness the source prior yields a large initial gain, but under structured block missingness it can \emph{fail entirely} and the learned flow becomes indispensable; the missingness-aware path contributes a further $12\%$--$26\%$ across all regimes. 
Mechanistically, structured priors shorten observed-region transport by up to $15\times$, concentrating the velocity field on genuinely uncertain entries. Project page: \url{https://zgp110.github.io/misspath-fm/}.
\end{abstract}

%% file: sec/1_intro.tex
\section{Introduction}
\label{sec:intro}

Real-world time series are rarely fully observed. In healthcare, laboratory tests are sparse and asynchronous; in energy systems, sensors fail or report irregularly; missing values often occur in structured blocks rather than isolated dropouts. Models must therefore operate under \emph{partial observability}, making predictions from incomplete and irregular evidence~\citep{che2018grud,rubanova2019latentode}.

Prior work addresses missingness through deterministic imputation, recurrent architectures, and probabilistic generative models.
Diffusion-based methods~\citep{tashiro2021csdi,alcaraz2022diffusion,Chen2023ProvablyCS} and transport-based approaches~\citep{chen2024newimp,clwf2024} have further advanced generative imputation.
While effective, these methods typically treat missingness as an input-side conditioning mechanism, overlooking how the \emph{geometry of generative transport} interacts with partial observability.

Flow matching has emerged as a powerful alternative to diffusion models~\citep{lipman2023flowmatching,liu2023rectifiedflow}, learning a continuous velocity field that transports a source distribution to the target via ODE integration~\citep{tong2023improving}.
Recent works apply flow matching to temporal data: TSFlow~\citep{kollovieh2025tsflow} integrates GP priors for forecasting, and TimeBridge~\citep{park2024timebridge} explores data-dependent priors for synthesis.
Despite this progress, \emph{none of these methods explicitly addresses the interaction between the flow-matching transport and partial observability}.
When $z_0 \sim \mathcal{N}(0,I)$, the model wastes capacity transporting already-observed entries from noise, complicating learning in genuinely uncertain regions.

We argue that the central bottleneck in applying flow matching to incomplete time series lies not in the velocity network architecture, but in the \emph{geometry of transport under partial observability}. Two questions naturally arise: (1)~what source distribution should the transport originate from when part of the sequence is already known? and (2)~how should the probability path allocate transport effort between observed and missing regions? Addressing these questions yields a simple design principle: the source prior should respect the observed signal, and the transport path should concentrate generative effort on the uncertain entries.

Motivated by this perspective, we propose \textbf{MissPath-FM} (\textbf{Miss}ingness-aware \textbf{Path} design for \textbf{F}low \textbf{M}atching), a partial-observation-aware flow matching framework for incomplete time series. MissPath-FM combines \emph{structured source priors} (from interpolation priors to Gaussian-process posterior priors) with a \emph{missingness-aware smooth path} and context conditioning based on missingness masks and time gaps. Instead of regenerating the entire sequence from pure noise, MissPath-FM learns a velocity field that transports uncertainty from observed evidence into the missing regions.

Empirically, our evaluation across 48 settings reveals that no single design choice is universally sufficient. Under random missingness (MCAR), the \emph{source prior} yields a large initial gain. Under structured block missingness, however, the prior alone can \emph{degrade} performance when contiguous gaps leave no flanking anchors; here the learned flow becomes indispensable. The missingness-aware path provides a further $12\%$--$26\%$ improvement across both regimes. Only the full pipeline---prior, flow, and path acting together---achieves robust performance across all missingness types. Mechanistically, structured priors shorten transport on observed regions by up to $15\times$, reallocating capacity toward genuinely uncertain entries.

Our primary contributions are:
\begin{itemize}[leftmargin=*, topsep=0pt, itemsep=0pt]
\item We introduce \textbf{MissPath-FM}, a flow matching framework for partial observability that combines structured source priors, mask/gap conditioning, and missingness-aware probability paths.
\item We establish a \textbf{regime-dependent design hierarchy} in which no single component suffices: the source prior yields a large initial gain under random missingness but can fail under block patterns; the learned flow is indispensable precisely where the prior struggles; and the missingness-aware path contributes a consistent $12\%$--$26\%$ across all regimes. Only their combination achieves robust performance.
\item We provide a \textbf{transport-geometric explanation}, showing how structured priors shorten observed-region trajectories and reallocate the velocity field's capacity to genuinely uncertain entries.
\item We demonstrate \textbf{state-of-the-art performance} across 32 settings at $T{=}48$ (4 datasets, 4 missingness mechanisms, 2 ratios), winning 29 against eight baselines (45/48 including $T \in \{24, 96\}$).
\end{itemize}

%% file: sec/2_related.tex
\section{Related Work}
\label{sec:related_work}

\paragraph{Time-series modeling under missingness and irregular sampling.}
A large body of literature studies multivariate time series with missing values and irregular observations. Early deep learning approaches such as GRU-D and BRITS incorporate missingness masks and time intervals into recurrent architectures, showing that missingness patterns themselves carry informative signals~\citep{che2018grud,cao2018brits}. Beyond recurrence, mTAN introduces learned continuous-time embeddings with multi-head temporal attention for irregularly sampled data~\citep{shukla2021mtan}, Neural CDE treats the time series as a control path driving latent dynamics~\citep{kidger2020neuralcde}, and SAITS replaces recurrence with masked self-attention for efficient deterministic imputation~\citep{du2023saits}. While effective, these methods target point estimation or classification and do not model the full conditional distribution over missing entries. MissPath-FM addresses this limitation by formulating partial observation as a \emph{conditional transport} problem, where the source prior, context encoding, and probability path are jointly designed around the missingness structure.

\paragraph{Probabilistic and generative models for incomplete time series.}
Another line of work tackles missingness through probabilistic and generative modeling. GP-VAE enforces a Gaussian-process prior in the latent space for smooth temporal structure~\citep{fortuin2020gpvae}, while Latent ODEs provide continuous-time latent dynamics that naturally accommodate non-uniform observations~\citep{rubanova2019latentode}. Within the diffusion framework, CSDI combines conditional score-based diffusion with a transformer backbone~\citep{tashiro2021csdi}; SSSD replaces the transformer with a structured state-space model, excelling in blackout-missing scenarios~\citep{alcaraz2022diffusion}; and PriSTI incorporates spatiotemporal dependencies as conditional priors~\citep{liu2023pristi}. Beyond standard diffusion, CSBI applies the Schr\"odinger bridge to time-series imputation via entropy-regularized optimal transport~\citep{Chen2023ProvablyCS}, and NewImp reinterprets diffusion-based imputation from a Wasserstein gradient flow perspective~\citep{chen2024newimp}. Despite strong empirical results, all these methods rely on predefined noise-corruption schedules or iterative bridge procedures, treating missingness primarily as an input-side conditioning signal. In contrast, MissPath-FM learns a continuous transport map whose source distribution is dynamically adapted to observed evidence, bypassing iterative denoising entirely.

\paragraph{Flow matching and transport-based generative modeling.}
Flow matching provides a simulation-free framework for training continuous normalizing flows by regressing velocity fields along predefined probability paths~\citep{lipman2023flowmatching}. Rectified-flow formulations~\citep{liu2023rectifiedflow} and OT-CFM with minibatch couplings~\citep{tong2023improving} further straighten trajectories for efficient few-step inference. These advances have catalyzed growing interest in transport-based models for temporal data: TFM extends flow matching to irregular clinical time series~\citep{zhang2024tfm}; FM-TS adapts rectified flows for generation~\citep{fmts2024}; TSFlow integrates GP priors for probabilistic forecasting~\citep{kollovieh2025tsflow}; CLWF uses Lagrangian OT flows for imputation~\citep{clwf2024}; PSW-I applies optimal transport with posterior sampling for imputation~\citep{wang2025optimal}; and TimeBridge explores data-dependent priors within a diffusion bridge~\citep{park2024timebridge}.
Three concurrent works are especially relevant. CFMI~\citep{simkus2025cfmi} applies conditional flow matching for general-purpose imputation but retains a Gaussian source without structured priors. CGFM~\citep{xu2025cgfm} shares our motivation of informed priors by using auxiliary predictions as the source, but targets fully observed forecasting without mask conditioning. Impute-MACFM~\citep{liu2025imputemacfm} introduces a mask-aware CFM objective for tabular data that penalizes velocity on observed entries, addressing partial observability through loss design rather than prior/path shaping. MissPath-FM is complementary: rather than relying on loss design, we reshape the source distribution so the velocity field \emph{naturally} concentrates on missing entries, and pair it with a missingness-aware path that allocates transport asymmetrically.

%% file: sec/3_method_nips.tex
\section{Method}
\label{sec:method}

\begin{figure}[t]
\centering
\includegraphics[width=\textwidth]{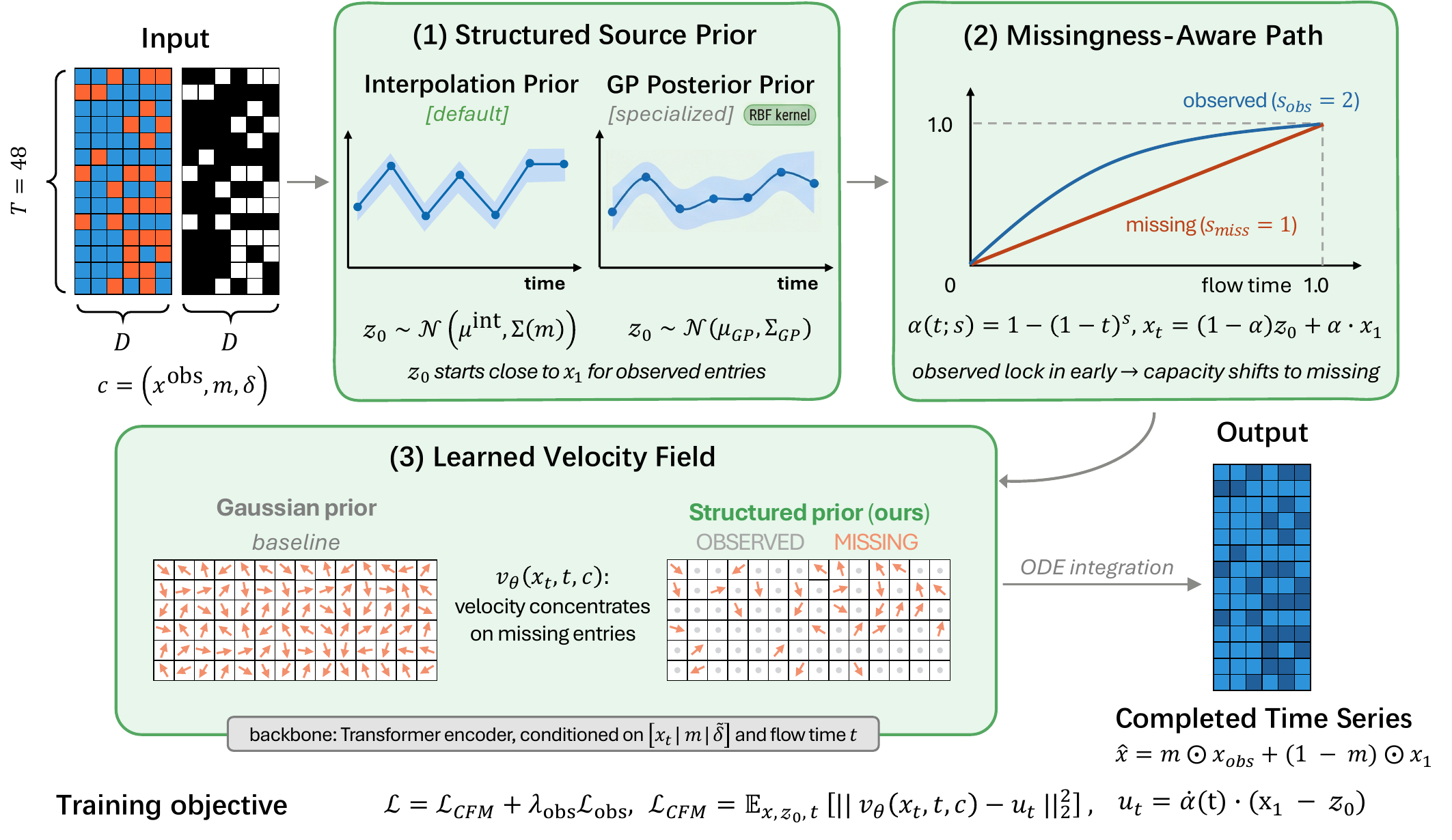}
\caption{Overview of MissPath-FM. \textbf{(1)}~A structured source prior initializes $z_0$ close to the observed signal. \textbf{(2)}~A missingness-aware path ($s_{\mathrm{obs}} > s_{\mathrm{miss}}$) locks in observed entries early. \textbf{(3)}~A context-conditioned Transformer predicts the velocity; at inference, $z_0$ is integrated with 20 Euler steps and observed entries are restored.}
\label{fig:framework}
\end{figure}

We present MissPath-FM, a generative framework for incomplete time series that reshapes the flow matching transport around partial observability (Figure~\ref{fig:framework}).
The method has three core components, all used in every reported result: (i)~an observation-aware source prior (Section~\ref{subsec:priors}), (ii)~a missingness-aware probability path (Section~\ref{subsec:path}), and (iii)~a context-conditioned velocity network (Section~\ref{subsec:optimization}).
Two further elements are not core: the GP posterior prior is an optional alternative to the default interpolation prior in (i) and appears only in ablations and analyses, and RandomRate is a training protocol rather than a part of the architecture (Section~\ref{sec:exp}).
Section~\ref{subsec:formulation} first identifies the geometric bottleneck that motivates all three; Table~\ref{tab:notation} collects the notation, and Figure~\ref{fig:pseudocode} gives the complete training step in code.

\begin{table}[!h]
\centering
\caption{Notation, grouped by role. The two source priors are alternative choices for the same object $p_0(\cdot\mid c)$, not sequential steps.}
\label{tab:notation}
\small
\setlength{\tabcolsep}{4pt}
\begin{tabular}{@{}ll@{\hspace{12pt}}ll@{}}
\toprule
\multicolumn{2}{@{}l}{\emph{Data and context}} & \multicolumn{2}{l@{}}{\emph{Source prior (choose one)}} \\
$x \in \mathbb{R}^{T\times D}$ & complete series & $\mu^{\mathrm{int}}, \Sigma(m)$ & Option A: interpolation (default) \\
$m \in \{0,1\}^{T\times D}$ & mask, $1$ = observed & $\mu_\star, \Sigma_\star$ & Option B: GP posterior (optional) \\
$\delta$ & time since last observation & $\sigma_{\mathrm{obs}}, \sigma_{\mathrm{miss}}$ & prior noise scales \\
$c = (x^{\mathrm{obs}}, m, \delta)$ & conditioning context & & \\
\midrule
\multicolumn{2}{@{}l}{\emph{Flow matching}} & \multicolumn{2}{l@{}}{\emph{Path schedule}} \\
$z_0 \sim p_0(\cdot \mid c)$ & source endpoint ($t{=}0$) & $s_{\mathrm{obs}}, s_{\mathrm{miss}}$ & transport speeds \\
$x_1 \sim p_{\mathrm{data}}$ & target endpoint ($t{=}1$) & $s_{\tau,d}$ & entry-wise speed (Eq.~\ref{eq:speed}) \\
$x_t$ & interpolant at flow time $t$ & $\alpha_{\tau,d}(t)$ & entry-wise schedule (Eq.~\ref{eq:path}) \\
$v_\theta$,\ $u_t$ & learned / target velocity & $\mathcal{C}_{\mathrm{obs}}, \mathcal{C}_{\mathrm{miss}}$ & transport cost (Prop.~\ref{prop:transport}) \\
\bottomrule
\end{tabular}
\end{table}

\subsection{Problem Formulation and Geometric Bottleneck}
\label{subsec:formulation}

Let $x \in \mathbb{R}^{T \times D}$ denote a multivariate time series of length $T$ with $D$ variables. A binary mask $m \in \{0,1\}^{T \times D}$ indicates observed entries ($m_{\tau,d}=1$), the observed input is $x^{\mathrm{obs}} = m \odot x$, and a time-gap tensor $\delta \in \mathbb{R}_{\ge 0}^{T \times D}$ records the elapsed time since the last observation of each variable. Our goal is to model $p(x \mid c)$ given context $c = (x^{\mathrm{obs}}, m, \delta)$.

Flow matching~\citep{lipman2023flowmatching} learns a velocity field $v_\theta: \mathbb{R}^{T \times D} \times [0,1] \to \mathbb{R}^{T \times D}$ transporting a source distribution $p_0$ to $p_1 \approx p_{\mathrm{data}}$. Conditional flow matching (CFM)~\citep{lipman2023flowmatching,tong2023improving} constructs conditional paths $p_t(x \mid z)$ anchored by pairs $z = (z_0, x_1)$ with $z_0 \sim p_0$, $x_1 \sim p_{\mathrm{data}}$, yielding the training objective:
\begin{equation}
\label{eq:cfm_objective}
\mathcal{L}_{\mathrm{CFM}} = \mathbb{E}_{t,\, x_1,\, z_0} \left[ \left\| v_{\theta}(x_t,t) - u_t(x_t \mid z_0, x_1) \right\|_2^2 \right].
\end{equation}
Following Tong et al.~\citep{tong2023improving}, we write $z_0$ rather than $x_0$ for the source endpoint so that it is not confused with the data variable $x$: the transport runs from $z_0$ at $t{=}0$ to $x_1$ at $t{=}1$, and $x_t$ is the point in between.
The standard linear path $x_t = (1-t)z_0 + t\,x_1$ yields a constant conditional velocity $u_t = x_1 - z_0$. Under this path, the expected transport cost decomposes by observation status:

\begin{proposition}[Transport cost decomposition]
\label{prop:transport}
Under the linear conditional path, the expected squared transport distance splits into an observed and a missing part:
\begin{equation}
\label{eq:cost_decomp}
\mathbb{E}\bigl[\|x_1 - z_0\|^2\bigr] = \mathcal{C}_{\mathrm{obs}} + \mathcal{C}_{\mathrm{miss}},
\qquad
\mathcal{C}_{\mathrm{obs}} = \sum_{m_{\tau,d}=1} \mathbb{E}\bigl[(x_{1,\tau,d} - z_{0,\tau,d})^2\bigr],
\end{equation}
and $\mathcal{C}_{\mathrm{miss}}$ is the same sum over missing entries ($m_{\tau,d}=0$).
Let $n_{\mathrm{obs}} = \sum_{\tau,d} m_{\tau,d}$. Two choices of the source $p_0$ give:
\begin{enumerate}[label=(\alph*), leftmargin=*, topsep=1pt, itemsep=1pt]
\item \emph{Gaussian source} $z_0 \sim \mathcal{N}(0, I)$, independent of $x_1$: each observed entry contributes $\mu^2 + \sigma^2 + 1$, where $\mu, \sigma$ are its marginal mean and standard deviation. For z-scored data this equals $2$, so $\mathcal{C}_{\mathrm{obs}} = 2\,n_{\mathrm{obs}}$, growing linearly with the number of \emph{observed} entries.
\item \emph{Interpolation source} (Eq.~\ref{eq:interp_prior}): observed entries satisfy $z_0 = x_1 + \epsilon$ with $\epsilon \sim \mathcal{N}(0, \sigma_{\mathrm{obs}}^2)$, so $\mathcal{C}_{\mathrm{obs}} = \sigma_{\mathrm{obs}}^2\, n_{\mathrm{obs}} \to 0$ as $\sigma_{\mathrm{obs}} \to 0$.
\end{enumerate}
\end{proposition}

The proof follows from linearity of expectation and standard Gaussian moments (Appendix~\ref{app:proofs}). Proposition~\ref{prop:transport} quantifies a fundamental inefficiency: for normalized data at $50\%$ missingness, approximately half the network's capacity is wasted on trivially known entries. MissPath-FM eliminates this waste through structured source priors (Section~\ref{subsec:priors}) and further concentrates capacity via a missingness-aware path (Section~\ref{subsec:path}).

\subsection{Observation-Aware Source Priors}
\label{subsec:priors}

We replace the uninformed Gaussian $z_0 \sim \mathcal{N}(0, I)$ with source distributions $z_0 \sim p_0(\cdot \mid c)$ conditioned on the observation context, guided by two principles: (i)~\emph{consistency}---observed entries in $z_0$ should be close to their known values, driving $\mathcal{C}_{\mathrm{obs}} \to 0$; and (ii)~\emph{temporal coherence}---missing entries should be initialized with temporally plausible values rather than independent noise, reducing $\mathcal{C}_{\mathrm{miss}}$ and easing the learning burden on the velocity network.
We consider two instantiations of $p_0(\cdot \mid c)$, used \emph{alternatively} rather than in sequence.

\textbf{Option A: interpolation prior (default).} For each variable $d$, we apply piecewise linear interpolation along the time axis: observed entries retain their values; missing entries are interpolated between flanking observations (with forward/backward fill at boundaries). Denoting this $\mu^{\mathrm{int}} = \mathcal{I}(x^{\mathrm{obs}}, m)$, the source distribution is:
\begin{equation}
\label{eq:interp_prior}
z_0 \sim \mathcal{N}\!\left(\mu^{\mathrm{int}},\, \Sigma(m)\right), \quad
\Sigma_{\tau,d}(m) = 
\begin{cases}
\sigma_{\mathrm{obs}}^2, & \text{if } m_{\tau,d}=1, \\
\sigma_{\mathrm{miss}}^2, & \text{if } m_{\tau,d}=0.
\end{cases}
\end{equation}
Setting $\sigma_{\mathrm{obs}} \ll \sigma_{\mathrm{miss}}$ satisfies the consistency principle: $\mathcal{C}_{\mathrm{obs}} = \sigma_{\mathrm{obs}}^2\, n_{\mathrm{obs}}$ (Proposition~\ref{prop:transport}). This prior has no fitted parameters and is robust, but provides only \emph{pointwise} coherence---under structured missingness (e.g., contiguous block gaps), interpolation degenerates into flat extrapolation, violating temporal coherence.

\textbf{Option B: Gaussian process posterior prior (optional).} To provide \emph{joint} temporal coherence, we introduce a GP posterior prior. For each variable $d$, given observed timestamps $\mathbf{t}_o$ and values $\mathbf{y}_o$, we compute the posterior predictive over the full grid $\mathbf{t}_\star = (1, \ldots, T)$:
\begin{align}
\label{eq:gp_mean}
\mu_\star &= K_{\star o}\,(K_{oo} + \sigma_n^2 I)^{-1}\,\mathbf{y}_o, \\
\label{eq:gp_cov}
\Sigma_\star &= K_{\star\star} - K_{\star o}\,(K_{oo} + \sigma_n^2 I)^{-1}\,K_{o\star},
\end{align}
where $K_{\star o} = k_\psi(\mathbf{t}_\star, \mathbf{t}_o)$ (RBF kernel by default). Sampling $z_{0,:,d} \sim \mathcal{N}(\mu_\star, \Sigma_\star)$ injects temporal correlations unavailable to local interpolation: the GP posterior mean smoothly regresses toward the global trend within gaps, and the full covariance encodes structured uncertainty that grows with distance from observed anchors. The two priors are complementary---interpolation is robust but degrades in long gaps; GP exploits temporal structure at the cost of kernel sensitivity (Appendix~\ref{app:hparam}). \emph{Choosing between them.} All main results use Option A. Option B is $23$--$39\times$ slower to construct (Appendix~\ref{app:rob_scale}), so we recommend it only for short, smooth series where its temporal structure pays for that cost.
Both options drive $\mathcal{C}_{\mathrm{obs}}$ toward zero; Option B additionally reduces $\mathcal{C}_{\mathrm{miss}}$ (Section~\ref{subsec:mechanism_analysis}).

\subsection{Missingness-Aware Probability Path}
\label{subsec:path}

Even with a structured prior that reduces $\mathcal{C}_{\mathrm{obs}}$, the standard linear path applies a uniform temporal schedule to all entries. This is suboptimal: observed entries start near their targets and need minimal refinement, while missing entries require the full transport capacity. We introduce a missingness-aware (MA) path with entry-wise speed:
\begin{equation}
\label{eq:speed}
s_{\tau,d} = s_{\mathrm{miss}} + (s_{\mathrm{obs}} - s_{\mathrm{miss}})\,m_{\tau,d},
\end{equation}
where $s_{\mathrm{obs}} \ge s_{\mathrm{miss}} \ge 1$. This defines $\alpha_{\tau,d}(t) = 1 - (1-t)^{s_{\tau,d}}$ and the conditional path:
\begin{equation}
\label{eq:path}
x_{t,\tau,d} = \bigl(1 - \alpha_{\tau,d}(t)\bigr)\,z_{0,\tau,d} + \alpha_{\tau,d}(t)\,x_{1,\tau,d}.
\end{equation}
The conditional target velocity is:
\begin{equation}
\label{eq:target_vel}
u_{t,\tau,d} = \dot{\alpha}_{\tau,d}(t)\,(x_{1,\tau,d} - z_{0,\tau,d}), \quad \dot{\alpha}_{\tau,d}(t) = s_{\tau,d}\,(1-t)^{s_{\tau,d}-1}.
\end{equation}
Since $\alpha(0) = 0$, $\alpha(1) = 1$, and $\alpha$ is monotonically increasing and $C^1$ for all $s \ge 1$, the path is valid within the CFM framework~\citep{lipman2023flowmatching}.

Under the standard path, even when $z_0 \approx x_1$ on observed entries, the network must keep $v_\theta \approx 0$ there \emph{across the entire interval} $t \in [0,1]$. With $s_{\mathrm{obs}} = 2$, observed entries complete $51\%$ of their transport by $t = 0.3$ while missing entries ($s_{\mathrm{miss}} = 1$) complete $30\%$; for $t > 0.5$ the observed residual is negligible, and the velocity field effectively reduces from $T \times D$ to $|\mathcal{M}|$ dimensions.
Unlike the prior, the MA path leaves $\mathcal{C}_{\mathrm{obs}}$ and $\mathcal{C}_{\mathrm{miss}}$ essentially unchanged; what it shortens is the part of the trajectory during which the network must track observed entries, so its benefit shows up in accuracy rather than in trajectory length (Appendix~\ref{app:transport}).

\subsection{Velocity Network and Optimization}
\label{subsec:optimization}

We parameterize $v_\theta(x_t, t, c)$ with a Transformer encoder along the temporal dimension. At each step $\tau$, the input token is $[x_{t,\tau,:}\,\|\,m_{\tau,:}\,\|\,\tilde{\delta}_{\tau,:}] \in \mathbb{R}^{3D}$ (where $\tilde{\delta} = \delta / (\delta_{\max} + 1)$), projected to $d_{\mathrm{model}}$ and augmented with learnable positional and sinusoidal flow-time embeddings. Full architectural details are in Appendix~\ref{app:impl_details}.

Given $z_0 \sim p_0(\cdot \mid c)$ and $t \sim \mathcal{U}[0,1]$, we construct $x_t$ via Eq.~\eqref{eq:path} and minimize:
\begin{equation}
\label{eq:loss}
\mathcal{L} = \mathcal{L}_{\mathrm{CFM}} + \lambda_{\mathrm{obs}}\,\mathcal{L}_{\mathrm{obs}},
\qquad
\mathcal{L}_{\mathrm{CFM}} = \mathbb{E}_{x,\, z_0,\, t} \left[ \left\| v_{\theta}(x_t, t, c) - u_t \right\|_2^2 \right],
\end{equation}
where $u_t$ is the conditional target velocity from Eq.~\eqref{eq:target_vel}, $\mathcal{L}_{\mathrm{CFM}}$ is computed over all entries, and $\mathcal{L}_{\mathrm{obs}}$ is the same squared error averaged over observed entries only, with $\lambda_{\mathrm{obs}} = 0.1$ in all reported results. The auxiliary term is light: the structured prior and MA path already make the observed-entry targets small (Proposition~\ref{prop:transport}), so it acts as a consistency safeguard rather than as the mechanism that concentrates the velocity field, unlike explicit velocity penalization~\citep{liu2025imputemacfm}.

At inference, we sample $z_0 \sim p_0(\cdot \mid c_{\mathrm{test}})$ and integrate via fixed-step Euler: $x_{t+\Delta t} = x_t + \Delta t\, v_\theta(x_t, t, c_{\mathrm{test}})$ with $\Delta t = 1/N_{\mathrm{step}}$. Observed entries are deterministically restored after integration. For probabilistic imputation, $K$ independent samples yield an ensemble for pointwise statistics. MissPath-FM extends to forecasting by constructing the prior exclusively from the history window (Appendix~\ref{app:forecast_mode}).
Figure~\ref{fig:pseudocode} gives the complete training step.

\begin{figure}[!h]
\begin{lstlisting}[escapeinside={(*}{*)}]
# (i) Interpolation prior, Eq. (*\ref{eq:interp_prior}*)
mu    = interp_1d(x_obs, mask)             # piecewise linear in time
sigma = where(mask, 0.1, 1.0)              # sigma_obs << sigma_miss
z0    = mu + sigma * randn_like(mu)

# (ii) Missingness-aware path, Eqs. (*\ref{eq:speed}*)-(*\ref{eq:target_vel}*)
s     = where(mask, 2.0, 1.0)              # s_obs = 2, s_miss = 1
alpha = 1 - (1 - t) ** s
x_t   = (1 - alpha) * z0 + alpha * x1
u_t   = s * (1 - t) ** (s - 1) * (x1 - z0)

# (iii) Loss, Eq. (*\ref{eq:loss}*): CFM on all entries + light observed-entry term
err   = (v_theta(x_t, t, ctx) - u_t) ** 2
loss  = err.mean() + 0.1 * (mask * err).sum() / mask.sum()
\end{lstlisting}
\vspace{-2mm}
\caption{The complete MissPath-FM training step. Variables follow Table~\ref{tab:notation}; \texttt{ctx} is the context $c$.}
\label{fig:pseudocode}
\end{figure}

%% file: sec/4_experiment_nips.tex
\begin{table*}[t]
\centering
\caption{Imputation results (MAE $\downarrow$, $T{=}48$). Best in \textbf{bold}, second best \underline{underlined}. MissPath-FM wins \textbf{29/32} settings. All methods share identical splits and masks. Baselines use official code where available; FlowTS, TSFlow, and CLWF include re-implemented components (Appendix~\ref{app:baselines}).}
\label{tab:main_imputation}
\vspace{-2mm}
\renewcommand{\arraystretch}{1.00}
\resizebox{\textwidth}{!}{%
\fontsize{10pt}{11.5pt}\selectfont
\begin{tabular}{ll r cccc cccc >{\columncolor{ourscolor}}c}
\toprule
& & & \multicolumn{4}{c}{\textsc{Transport-based}} & \multicolumn{4}{c}{\textsc{Deterministic / Generative}} & \\
\cmidrule(lr){4-7} \cmidrule(lr){8-11}
Dataset & Mechanism & Rate & FlowTS & CLWF & TSFlow & PSW-I & SAITS & BRITS & CSDI & GP-VAE & \textbf{Ours} \\
\midrule
\multirow{8}{*}{\rotatebox[origin=c]{90}{\textbf{ETTm1}}}
  & MCAR      & 30\% & 0.393 & 0.190 & 0.121 & 0.144 & \underline{0.096} & 0.103 & 0.159 & 0.454 & \textbf{0.075} \\
  & MCAR      & 50\% & 0.431 & 0.224 & 0.132 & 0.162 & \underline{0.109} & 0.122 & 0.203 & 0.583 & \textbf{0.085} \\
  & Block     & 30\% & 0.396 & 0.197 & 0.245 & 0.265 & \underline{0.152} & 0.160 & 0.248 & 0.511 & \textbf{0.106} \\
  & Block     & 50\% & 0.450 & 0.249 & 0.267 & 0.320 & \underline{0.199} & 0.238 & 0.358 & 0.572 & \textbf{0.145} \\
  & MNAR      & 30\% & 0.439 & 0.366 & 0.186 & 0.243 & \underline{0.143} & 0.186 & 0.219 & 0.746 & \textbf{0.117} \\
  & MNAR      & 50\% & 0.482 & 0.382 & 0.244 & 0.285 & \underline{0.216} & 0.269 & 0.273 & 0.793 & \textbf{0.158} \\
  & Irregular & 30\% & 0.444 & 0.382 & 0.144 & 0.188 & \underline{0.134} & 0.141 & 0.246 & 0.735 & \textbf{0.101} \\
  & Irregular & 50\% & 0.461 & 0.327 & 0.141 & 0.188 & \underline{0.132} & 0.145 & 0.532 & 0.731 & \textbf{0.101} \\
\midrule
\multirow{8}{*}{\rotatebox[origin=c]{90}{\textbf{Synthetic}}}
  & MCAR      & 30\% & 0.702 & 0.579 & \underline{0.112} & 0.216 & 0.139 & 0.161 & 0.200 & 0.785 & \textbf{0.098} \\
  & MCAR      & 50\% & 0.794 & 0.729 & \underline{0.138} & 0.289 & 0.211 & 0.200 & 1.052 & 0.799 & \textbf{0.110} \\
  & Block     & 30\% & 0.791 & 0.761 & 0.555 & 0.821 & 0.593 & \underline{0.547} & 0.688 & 0.778 & \textbf{0.374} \\
  & Block     & 50\% & 0.823 & 0.784 & 0.645 & 0.893 & \underline{0.641} & 0.794 & 0.726 & 0.783 & \textbf{0.470} \\
  & MNAR      & 30\% & 0.866 & 0.841 & \underline{0.119} & 0.308 & 0.143 & 0.193 & 1.106 & 0.994 & \textbf{0.115} \\
  & MNAR      & 50\% & 0.918 & 0.873 & \underline{0.141} & 0.401 & 0.256 & 0.246 & 1.038 & 0.924 & \textbf{0.126} \\
  & Irregular & 30\% & 0.841 & 0.804 & \underline{0.248} & 0.427 & 0.377 & 0.340 & 0.993 & 0.801 & \textbf{0.199} \\
  & Irregular & 50\% & 0.843 & 0.811 & \underline{0.245} & 0.425 & 0.379 & 0.335 & 1.031 & 0.800 & \textbf{0.196} \\
\midrule
\multirow{8}{*}{\rotatebox[origin=c]{90}{\textbf{Electricity}}}
  & MCAR      & 30\% & 0.309 & 0.232 & 0.178 & 0.155 & 0.159 & \underline{0.145} & 0.274 & 0.336 & \textbf{0.091} \\
  & MCAR      & 50\% & 0.324 & 0.231 & 0.193 & 0.183 & 0.170 & \underline{0.169} & 0.330 & 0.359 & \textbf{0.100} \\
  & Block     & 30\% & 0.344 & 0.263 & 0.366 & 0.498 & \underline{0.184} & \textbf{0.178} & 0.309 & 0.361 & 0.216 \\
  & Block     & 50\% & 0.361 & 0.258 & 0.355 & 0.531 & \textbf{0.195} & \underline{0.200} & 0.375 & 0.417 & 0.252 \\
  & MNAR      & 30\% & 0.324 & 0.247 & 0.208 & 0.199 & 0.181 & \underline{0.170} & 0.333 & 0.487 & \textbf{0.112} \\
  & MNAR      & 50\% & 0.422 & 0.310 & 0.255 & 0.289 & \underline{0.213} & 0.216 & 0.369 & 0.598 & \textbf{0.130} \\
  & Irregular & 30\% & 0.416 & 0.365 & 0.275 & 0.236 & \underline{0.189} & 0.197 & 0.723 & 0.425 & \textbf{0.118} \\
  & Irregular & 50\% & 0.409 & 0.306 & 0.239 & 0.237 & \underline{0.186} & 0.201 & 1.070 & 0.409 & \textbf{0.120} \\
\midrule
\multirow{8}{*}{\rotatebox[origin=c]{90}{\textbf{PhysioNet}}}
  & Art.      & 10\% & 0.524 & 0.334 & 0.399 & 0.379 & \underline{0.234} & 0.257 & 0.363 & 0.594 & \textbf{0.227} \\
  & Art.      & 20\% & 0.558 & 0.346 & 0.407 & 0.391 & \underline{0.243} & 0.279 & 0.394 & 0.616 & \textbf{0.238} \\
  & Art.      & 30\% & 0.523 & 0.364 & 0.413 & 0.400 & \underline{0.264} & 0.299 & 0.443 & 0.638 & \textbf{0.250} \\
  & Art.      & 50\% & 0.545 & 0.412 & 0.440 & 0.427 & \underline{0.313} & 0.347 & 0.571 & 0.687 & \textbf{0.308} \\
  & Block     & 30\% & 0.550 & 0.358 & 0.407 & 0.400 & \underline{0.264} & 0.300 & 0.451 & 0.643 & \textbf{0.261} \\
  & Block     & 50\% & 0.536 & 0.417 & 0.435 & 0.427 & \underline{0.312} & 0.345 & 0.558 & 0.687 & \textbf{0.299} \\
  & MNAR      & 30\% & 0.524 & 0.363 & 0.419 & 0.399 & \underline{0.266} & 0.300 & 0.420 & 0.644 & \textbf{0.261} \\
  & MNAR      & 50\% & 0.540 & 0.410 & 0.438 & 0.425 & \textbf{0.314} & 0.346 & 0.514 & 0.687 & \underline{0.327} \\
\bottomrule
\end{tabular}%
}
\end{table*}

\section{Experimental Setup}
\label{sec:exp}

\paragraph{Datasets and evaluation protocol.}
We evaluate on four datasets:
\textbf{Synthetic} ($D{=}8$; multivariate sinusoidal),
\textbf{ETTm1} ($D{=}7$; electricity transformer temperatures)~\citep{zhou2021informer},
\textbf{Electricity} ($D{=}20$; UCI Electricity, top-variance clients)~\citep{dua2017uci}, and
\textbf{PhysioNet 2012} ($D{=}36$; ICU variables, ${\sim}19\%$ natural observation rate)~\citep{silva2012physionet}.
The default sequence length is $T{=}48$; we additionally test $T \in \{24, 96\}$ on ETTm1.
We assess three tasks: imputation, forecasting ($h{=}24$), and unconditional generation.
Missingness is imposed under four mechanisms at two ratios (30\%, 50\%): MCAR, Block (contiguous per-feature gaps), MNAR (value-dependent dropout), and Irregular (asynchronous per-feature patterns), yielding 32 settings at $T{=}48$.
PhysioNet retains its natural sparsity with 10\%--50\% additional artificial holdout for evaluation.
All features are z-score normalized per variable; data is split 80/20 with fixed seed 42.
Full specifications are in Appendix~\ref{app:dataset_details}.

\paragraph{Baselines and metrics.}
We compare against eight methods under a unified protocol (identical splits, masks, metrics):
SAITS~\citep{du2023saits} and BRITS~\citep{cao2018brits} (deterministic);
CSDI~\citep{tashiro2021csdi} and GP-VAE~\citep{fortuin2020gpvae} (diffusion/VAE);
FlowTS~\citep{fmts2024}, CLWF~\citep{clwf2024}, TSFlow~\citep{kollovieh2025tsflow}, and PSW-I~\citep{wang2025optimal} (transport-based).
Baselines use official codebases where available; when official code does not cover a task, we provide faithful re-implementations (Appendix~\ref{app:baselines}).
We report \textbf{MAE} on held-out missing positions for imputation and forecasting; for generation, \textbf{discriminative score}, \textbf{predictive score}, and \textbf{spectral distance} following~\cite{yoon2019timegan} (definitions in Appendix~\ref{app:gen_metrics}).
Our model uses a 4-layer Transformer encoder ($d_{\mathrm{model}}{=}256$, 4 heads, ${\sim}3.9$M parameters) with the interpolation prior and RandomRate training ($r \sim \mathcal{U}(0.1, 0.7)$ per sample), trained for 500 epochs with AdamW and Euler-20 inference. Full details are in Appendix~\ref{app:impl_details}.

\paragraph{Hyperparameter protocol.}
A single global configuration is used for every dataset, missingness mechanism, rate, and sequence length: the architecture above, interpolation prior with $\sigma_{\mathrm{obs}}{=}0.1$ and $\sigma_{\mathrm{miss}}{=}1.0$, MA path with $s_{\mathrm{obs}}{=}2$ and $s_{\mathrm{miss}}{=}1$, $\lambda_{\mathrm{obs}}{=}0.1$, RandomRate training, Euler-20, and $K{=}20$ samples.
No hyperparameter is tuned per dataset, per mechanism, or per rate.
Appendix~\ref{app:hparam} varies one axis at a time around this fixed configuration.

%% file: sec/5_results_nips.tex
\section{Results and Analysis}
\label{sec:results}

\subsection{Main Imputation Results}
\label{subsec:main_imputation_results}
 
Table~\ref{tab:main_imputation} reports MAE across all 32 evaluation settings (4 datasets $\times$ 4 missingness mechanisms $\times$ 2 ratios) at $T{=}48$.
MissPath-FM wins \textbf{29 of 32 settings} with an average rank of 1.16 across all methods.
The three losses occur in two specific regimes: Electricity Block 30\% and 50\% (BRITS and SAITS, whose recurrent and attention-based architectures handle moderate contiguous gaps on high-dimensional data effectively) and PhysioNet MNAR 50\% (SAITS, by $4\%$).
 
\begin{figure*}[t]
\centering
\includegraphics[width=\textwidth]{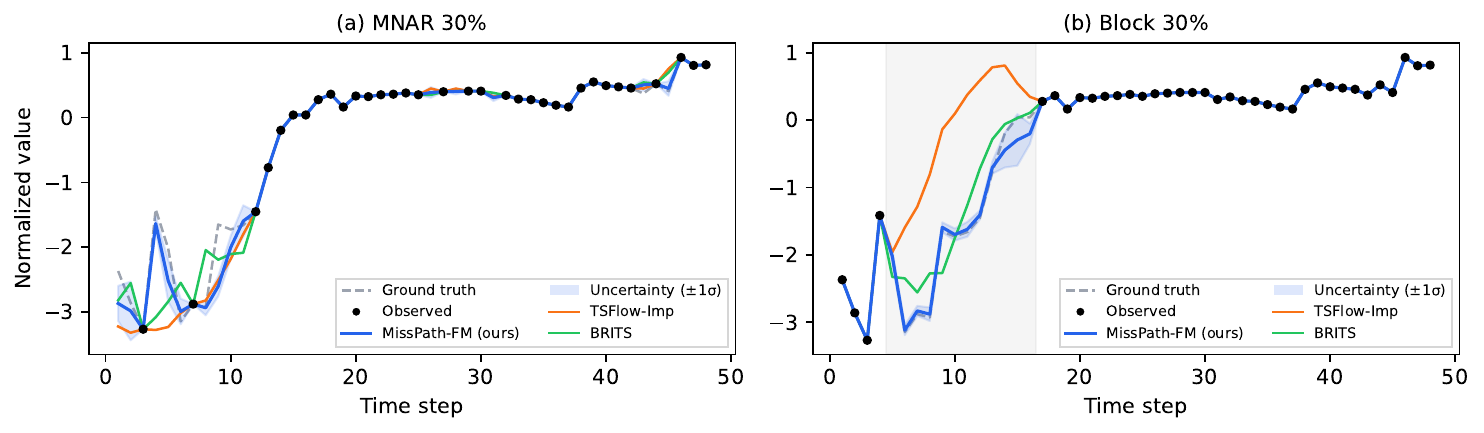}
\vspace{-4mm}
\caption{Qualitative imputation on ETTm1. 
\textbf{(a)}~MNAR 30\%: MissPath-FM recovers masked extreme 
values where baselines deviate. \textbf{(b)}~Block 30\%: within 
the contiguous gap (shaded), TSFlow diverges and BRITS 
flattens, while MissPath-FM tracks the ground truth with 
calibrated uncertainty.}
\label{fig:qualitative}
\end{figure*}

MissPath-FM's advantage is consistent across missingness mechanisms: it wins all 10 MCAR-type settings (including PhysioNet's artificial holdout), 6/8 Block, 7/8 MNAR, and 6/6 Irregular, confirming that observation-aware transport design generalizes beyond any single missingness type.
The margins are largest under structured missingness (ETTm1 Block 50\%: $-27\%$; Electricity MNAR 30\%: $-34\%$), where the geometric advantage of structured priors is most pronounced.
On PhysioNet, MissPath-FM wins 7 of 8 settings with narrow but consistent margins over SAITS, the strongest baseline on clinical data.
Figure~\ref{fig:qualitative} illustrates two representative cases: under MNAR, MissPath-FM recovers masked extreme values where baselines deviate; under Block, it tracks the ground truth within contiguous gaps with calibrated uncertainty.
We additionally evaluate at $T \in \{24, 96\}$ on ETTm1 (Appendix~\ref{app:t_variation}): MissPath-FM wins all 16 settings, with margins growing at longer sequences ($19\%$ at $T{=}24$, $26\%$ at $T{=}96$).
Beyond point estimates, \textbf{probabilistic calibration (Appendix~\ref{app:calibration})} and \textbf{multi-seed evaluation (Appendix~\ref{app:multiseed})} confirm both well-calibrated uncertainty (29/32 wins, ACE $\le 0.104$) and low variance (CV $< 2.4\%$).
Appendix~\ref{app:robustness} further reports a 10--90\% missing-rate sweep, an observation-noise study, an evaluation at $T{=}192$, and baselines retrained with the same RandomRate augmentation; the last shows that the augmentation does not help the baselines, so our margins do not stem from it.

\begin{table*}[t]
\vspace{-2mm}
\centering
\caption{Incremental ablation (MAE $\downarrow$, 4 d.p.). Configurations A--F add one component at a time. Best per row in \textbf{bold}. $\Delta_\mathrm{flow}$: gain of (F) over best Prior-Only.}
\vspace{-2mm}
\label{tab:ablation}
\renewcommand{\arraystretch}{1.08}
\resizebox{0.9\textwidth}{!}{%
\fontsize{10pt}{10pt}\selectfont
\begin{tabular}{ll c cc ccc >{\columncolor{ourscolor}}c}
\toprule
& & & \multicolumn{2}{c}{\textsc{Prior-Only}} & \multicolumn{3}{c}{\textsc{Incremental Flow}} & \\
\cmidrule(lr){4-5} \cmidrule(lr){6-8}
Dataset & Setting
  & \shortstack[c]{(A)\\Gaussian\\+Flow}
  & \shortstack[c]{(B)\\Interpolation\\Prior}
  & \shortstack[c]{(C)\\GP\\Prior}
  & \shortstack[c]{(D)\\+Flow\\Matching}
  & \shortstack[c]{(E)\\+Context\\Encoding}
  & \shortstack[c]{(F)\\+Missingness-\\Aware Path}
  & $\Delta_\mathrm{flow}$ \\
\midrule
\multirow{2}{*}{ETTm1}
  & MCAR 30\% & 0.7387 & 0.1091 & 0.1214 & 0.0918 & 0.0856 & \textbf{0.0751} & $-31\%$ \\
  & Block 30\% & 0.7248 & 0.2162 & 0.4150 & 0.1335 & 0.1253 & \textbf{0.1057} & $-51\%$ \\
\midrule
\multirow{2}{*}{Synthetic}
  & MCAR 30\% & 0.8165 & 0.1614 & 0.1089 & 0.1208 & 0.1182 & \textbf{0.0981} & $-10\%$ \\
  & Block 30\% & 0.8056 & 0.8574 & 0.6326 & 0.5578 & 0.4938 & \textbf{0.3737} & $-41\%$ \\
\midrule
\multirow{2}{*}{Electricity}
  & MCAR 30\% & 0.4028 & 0.1897 & 0.2230 & 0.1238 & 0.1137 & \textbf{0.0910} & $-52\%$ \\
  & Block 30\% & 0.4142 & 0.8417 & 0.7395 & 0.2789 & 0.2503 & \textbf{0.2158} & $-71\%$ \\
\midrule
\multirow{2}{*}{PhysioNet}
  & Art.\ 10\% & 0.6355 & 0.3632 & 0.4119 & 0.3305 & 0.3057 & \textbf{0.2271} & $-37\%$ \\
  & Block 30\% & 0.6307 & 0.3737 & 0.4404 & 0.3420 & 0.3248 & \textbf{0.2610} & $-30\%$ \\
\bottomrule
\end{tabular}%
}
\vspace{-1mm}
\end{table*}

\subsection{Ablation: What Drives the Gains?}
\label{subsec:ablation}
 
Table~\ref{tab:ablation} reports an incremental ablation across all four datasets (MCAR 30\% + Block 30\%), \textbf{adding one component at a time}.
Full per-component details are in Appendix~\ref{app:ablations}.
A notable property is that \emph{every} incremental component improves MAE in \emph{every} setting: (B)$\to$(D), (D)$\to$(E), and (E)$\to$(F) are all strictly negative across all 8 rows, confirming that each design choice contributes independently.
 
\paragraph{MCAR: prior provides the dominant initial gain.}
Under random missingness, replacing the Gaussian with interpolation reduces MAE by $43\%$--$85\%$, the single largest step.
The learned flow adds a further $9\%$--$35\%$ over the prior alone ((B)$\to$(D)).
The MA path contributes a consistent $12\%$--$26\%$ ((E)$\to$(F)), indicating that missingness-aware scheduling is beneficial even under random patterns, where the obs/miss asymmetry is sufficient for differential transport to improve accuracy.
 
\paragraph{Block: flow provides the dominant recovery.}
The pattern reverses under block missingness: interpolation-only \emph{fails} on Electricity (MAE $= 0.842$, worse than Gaussian+Flow at $0.414$) and is near-random on Synthetic (MAE $= 0.857$), because contiguous gaps leave no flanking observations for interpolation.
Here the learned flow provides the dominant recovery ($-67\%$ on Electricity Block, $-35\%$ on Synthetic Block via (B)$\to$(D)).
The MA path contributes $14\%$--$24\%$ in these hard regimes, consistent with its design purpose of concentrating capacity on uncertain entries.
 
\paragraph{Summary.}
Under MCAR the prior dominates ($-65\%$ on average) and the flow is secondary ($-21\%$); under block missingness the prior can be harmful and the flow becomes the primary driver ($-37\%$); the MA path contributes a consistent $18\%$--$19\%$ in both regimes.
A complementary leave-one-out ablation, removing one component from the full model at a time (Appendix~\ref{app:rob_loo}), confirms that every component remains necessary: removing the prior raises MAE by $373\%$ on average, context encoding by $23\%$, the MA path by $20\%$, and RandomRate by $18\%$.

\begin{table*}[t]
\centering
\caption{Forecasting results at $h{=}24$ (MAE $\downarrow$). Best in \textbf{bold}, second \underline{underlined}.}
\label{tab:forecasting_results}
\vspace{-2mm}
\setlength{\tabcolsep}{3.5pt}
\renewcommand{\arraystretch}{1.05}
\fontsize{7.5pt}{9.5pt}\selectfont
\begin{tabular}{ll ccccccc >{\columncolor{ourscolor}}c}
\toprule
Dataset & History & SAITS & BRITS & CSDI & GP-VAE & FlowTS & CLWF & TSFlow & \textbf{Ours} \\
\midrule
\multirow{2}{*}{ETTm1}
  & Clean     & 0.489 & 0.692 & 0.763 & 0.694 & 0.702 & 0.483 & \underline{0.355} & \textbf{0.266} \\
  & 30\% MCAR & 0.528 & 0.692 & 0.593 & 0.704 & 0.679 & 0.480 & \underline{0.365} & \textbf{0.264} \\
\midrule
\multirow{2}{*}{Synthetic}
  & Clean     & 0.791 & 0.781 & 1.084 & 0.787 & 0.850 & 0.825 & \underline{0.566} & \textbf{0.511} \\
  & 30\% MCAR & 0.806 & 0.778 & 1.071 & 0.794 & 0.850 & 0.837 & \underline{0.573} & \textbf{0.572} \\
\midrule
\multirow{2}{*}{Electricity}
  & Clean     & 1.040 & 0.873 & 0.861 & 0.937 & 0.935 & 0.947 & \underline{0.226} & \textbf{0.149} \\
  & 30\% MCAR & 0.871 & 0.821 & 1.056 & 0.907 & 0.918 & 0.953 & \underline{0.237} & \textbf{0.153} \\
\bottomrule
\end{tabular}
\vspace{-1mm}
\end{table*}

\begin{table*}[t]
\vspace{-1mm}
\centering
\caption{Generation quality (all metrics $\downarrow$). Best in \textbf{bold}, second \underline{underlined}.}
\label{tab:generation_results}
\vspace{-2mm}
\setlength{\tabcolsep}{5pt}
\renewcommand{\arraystretch}{1.05}
\fontsize{8pt}{10pt}\selectfont
\begin{tabular}{l ccc ccc ccc}
\toprule
& \multicolumn{3}{c}{Synthetic} & \multicolumn{3}{c}{ETTm1} & \multicolumn{3}{c}{Electricity} \\
\cmidrule(lr){2-4} \cmidrule(lr){5-7} \cmidrule(lr){8-10}
Method & Disc. & Pred. & Spec. & Disc. & Pred. & Spec. & Disc. & Pred. & Spec. \\
\midrule
CSDI       & 0.500 & 0.925 & 1.540 & 0.474 & 0.745 & 1.313 & 0.500 & 0.631 & 1.294 \\
GP-VAE     & 0.500 & 0.809 & 1.657 & 0.500 & 0.887 & 1.062 & 0.477 & 0.581 & 1.341 \\
FlowTS     & \underline{0.275} & 0.266 & \underline{0.126} & \underline{0.184} & \underline{0.214} & \underline{0.045} & 0.250 & 0.279 & \underline{0.112} \\
TSFlow     & 0.500 & 0.929 & 1.538 & 0.500 & 0.906 & 1.699 & 0.500 & 0.734 & 1.655 \\
Gauss.+Std & 0.300 & \textbf{0.240} & 0.155 & 0.198 & 0.286 & 0.256 & \underline{0.182} & \underline{0.201} & 0.113 \\
\rowcolor{ourscolor}
\textbf{MissPath-FM} & \textbf{0.169} & \underline{0.261} & \textbf{0.092} & \textbf{0.064} & \textbf{0.186} & \textbf{0.008} & \textbf{0.155} & \textbf{0.141} & \textbf{0.104} \\
\bottomrule
\end{tabular}
\vspace{-3mm}
\end{table*}

\subsection{Forecasting and Generation}
\label{subsec:beyond_imputation}
 
\paragraph{Forecasting.}
Table~\ref{tab:forecasting_results} reports forecasting at $h{=}24$ across three datasets with clean and corrupted histories.
MissPath-FM wins all 6 settings, with margins of $25\%$--$35\%$ on ETTm1 and Electricity; the Synthetic MCAR margin is narrow ($0.572$ vs $0.573$), consistent with TSFlow's strength on smooth signals.
On ETTm1 and Electricity the margin widens under history corruption, consistent with observation-aware transport being most useful when histories are partial.

\paragraph{Generation.}
Table~\ref{tab:generation_results} evaluates unconditional generation across three datasets.
MissPath-FM wins 8 of 9 metric--dataset combinations, including the best \emph{discriminative score} on all three datasets ($0.064$--$0.169$), indicating that generated samples are nearly indistinguishable from real data.
The single loss is the predictive score on Synthetic ($0.261$ vs Gauss+Std $0.240$), where the uninformed Gaussian flow slightly better preserves next-step dynamics on smooth sinusoidal signals.
Since all entries are missing in the unconditional setting, the generation comparison isolates the benefit of training-time transport efficiency (Appendix~\ref{app:gen_protocol}).

\subsection{Transport Geometry Analysis}
\label{subsec:mechanism_analysis}

\begin{figure*}[t]
\vspace{-1mm}
\centering
\begin{minipage}[b]{0.51\textwidth}
\centering
\includegraphics[width=\textwidth]{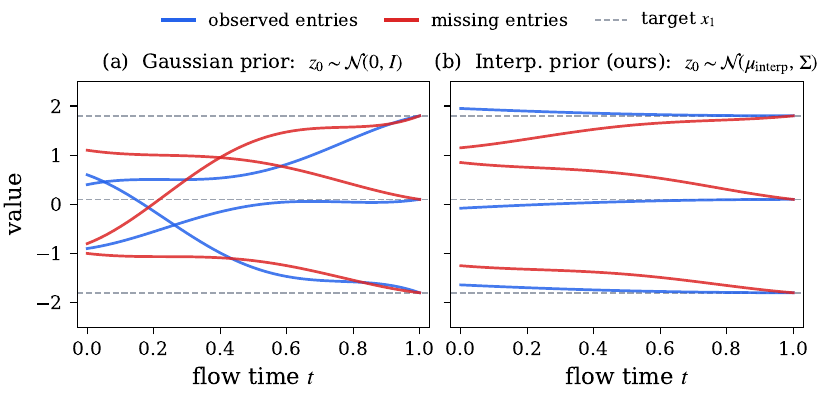}
\captionof{figure}{Transport trajectories under Gaussian prior 
\textbf{(a)} and interpolation prior \textbf{(b)}.}
\label{fig:trajectory}
\end{minipage}%
\hfill
\begin{minipage}[b]{0.48\textwidth}
\centering
\setlength{\tabcolsep}{4.5pt}
\fontsize{9pt}{10pt}\selectfont
\begin{tabular}{lcccc}
\toprule
Config & MAE & $\ell_{\mathrm{obs}}$ & $\ell_{\mathrm{miss}}$ & Ratio \\
\midrule
Gauss.+Std  & 0.882 & 0.716 & 0.757 & 1.1$\times$ \\
Interp+Std  & 0.651 & 0.047 & 0.906 & 19.2$\times$ \\
Interp+MA   & 0.634 & 0.048 & 0.895 & 18.9$\times$ \\
GP+Std      & 0.558 & 0.043 & 0.557 & 13.0$\times$ \\
GP+MA       & \textbf{0.534} & 0.042 & 0.552 & 13.1$\times$ \\
\bottomrule
\end{tabular}
\captionof{table}{Trajectory-length analysis (Synth.\ Block 50\%). $\ell_{\mathrm{obs}}$/$\ell_{\mathrm{miss}}$: mean arc length on observed/missing entries.}
\label{tab:trajectory_analysis}
\end{minipage}
\vspace{-1mm}
\end{figure*}

\begin{figure*}[t]
\vspace{-2mm}
\centering
\includegraphics[width=0.92\textwidth]{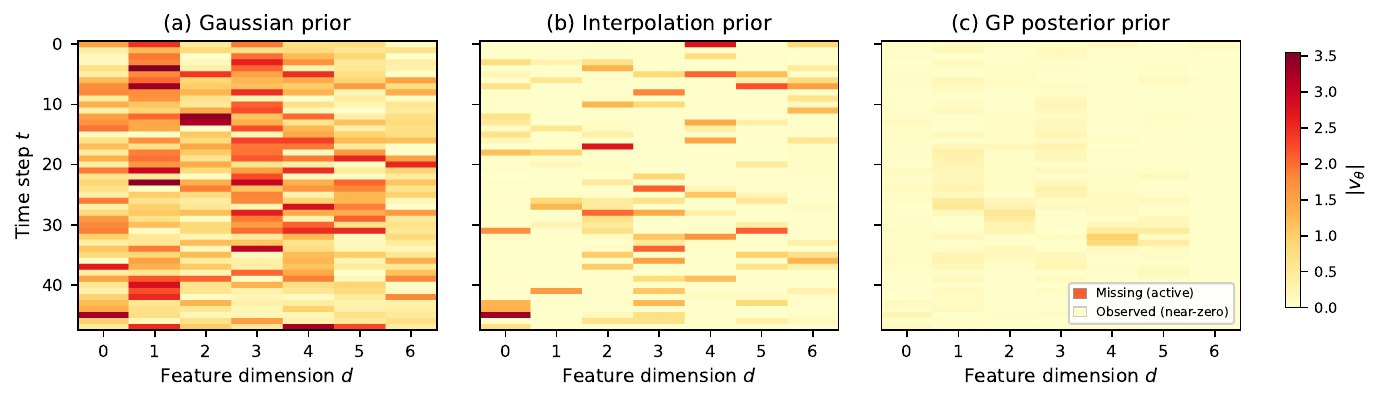}
\vspace{-2mm}
\caption{Velocity field magnitude $\|v_\theta\|$ at $t{=}0.5$ on ETTm1 MCAR 30\%. \textbf{(a)}~Gaussian prior: velocity is spread across all entries. \textbf{(b)}~Interpolation prior: observed entries are silenced and velocity concentrates on missing entries. \textbf{(c)}~GP posterior prior: missing-entry velocity is further reduced, since the posterior mean starts closer to the target.}
\label{fig:velocity_field}
\vspace{-5mm}
\end{figure*}

Figure~\ref{fig:trajectory} and Table~\ref{tab:trajectory_analysis} 
directly quantify the transport reallocation on Synthetic Block 50\%.
Under the Gaussian prior, all entries travel long distances from 
noise to target; under the interpolation prior, observed entries 
stay near targets (${\sim}15\times$ shorter transport).
The miss-to-obs ratio increases from $1.1\times$ to $19.2\times$, 
confirming that the velocity field concentrates on genuinely 
uncertain entries.
The GP prior additionally shortens missing-region transport by 
$39\%$, explaining its advantage under block and MNAR settings.
Figure~\ref{fig:velocity_field} visualizes this progression at 
the entry level: from uniform velocity under the Gaussian 
prior~(a), to concentrated missing-entry velocity under 
interpolation~(b), to further reduced velocity under the GP 
prior~(c).
The same effect is visible in the final samples: measured \emph{before} the deterministic restoration step, observed-entry MAE is $0.023$--$0.025$ for MissPath-FM against $0.40$--$0.77$ for a Gaussian-source flow (Appendix~\ref{app:rob_consistency}), so observed values are held in place by the transport itself rather than by the final projection.
This perspective explains why many alternatives fail: once the 
prior removes the dominant inefficiency, regularizers, cost-aware 
paths, and larger backbones have little residual room to 
contribute (Appendix~\ref{app:negative},~\ref{app:hparam}).

%% file: sec/6_conclusion.tex
\section{Conclusion}
\label{sec:conclusion}

We presented \textbf{MissPath-FM}, a flow matching framework that treats partial observability as a transport design problem rather than an input-side conditioning mechanism.
By combining structured source priors, a missingness-aware probability path, and context-conditioned velocity estimation, MissPath-FM reshapes the generative transport to concentrate on genuinely uncertain entries.
MissPath-FM wins 45 of 48 evaluation settings (93.8\%) across four datasets, four missingness mechanisms, and three sequence lengths against eight baselines.

\paragraph{Limitations and future work.}
Our largest tested setting is $D{=}36$ and $T{=}192$; behaviour at larger scale is not established.
The advantage shrinks when observations are very sparse or unreliable: against SAITS, MissPath-FM falls behind on two of four rate-sweep settings at $90\%$ missingness, and under observation noise beyond about $0.2$ (ETTm1) to $0.3$ (Electricity) z-scored units (Appendix~\ref{app:robustness}).
The GP prior is per-variable and ignores cross-variable dependencies; a multi-output GP could exploit them, but its cost grows cubically in $D$, already the bottleneck at $D{=}36$.
Downstream decision tasks and continuous-time irregular settings remain open.
More broadly, we hope this work encourages treating partial observability as a \emph{transport design problem} for generative time-series modeling.

%% file: sec/7_appendix.tex
\section*{Appendix Overview}
\begin{itemize}[leftmargin=*, itemsep=1pt, topsep=2pt]
\item[\ref{app:dataset_details}.] Dataset and Missingness Details
\item[\ref{app:impl_details}.] Implementation Details
\item[\ref{app:baselines}.] Baselines and Evaluation Metrics
\item[\ref{app:proofs}.] Proofs and Additional Method Details
\item[\ref{app:t_variation}.] Sequence Length Generalization
\item[\ref{app:calibration}.] Uncertainty Calibration
\item[\ref{app:ablations}.] Ablation Details
\item[\ref{app:hparam}.] Hyperparameter Sensitivity
\item[\ref{app:extended}.] Extended Analysis (Transport Geometry, Gap-Bucketed)
\item[\ref{app:multiseed}.] Multi-Seed Evaluation
\item[\ref{app:robustness}.] Robustness and Additional Analyses (Leave-One-Out, Consistency, Fairness, Rate, Noise, Scale)
\item[\ref{app:negative}.] Negative Results
\item[\ref{app:repro}.] Reproducibility
\end{itemize}

\section{Dataset and Missingness Details}
\label{app:dataset_details}
\label{app:missingness_details}

Table~\ref{tab:dataset_summary} summarizes the four evaluation datasets, which span three application domains (synthetic, energy, clinical) and a wide range of dimensionalities ($D \in \{7, 8, 20, 36\}$) and natural sparsity levels.
All datasets use non-overlapping windows, an 80/20 train/validation split with random seed 42, and per-variable z-score normalization.
PhysioNet retains its natural clinical missingness (${\sim}19\%$ observation rate); for the remaining three datasets, missingness is synthetically imposed to enable controlled evaluation.

\begin{table*}[h]
\centering
\caption{Dataset summary.}
\label{tab:dataset_summary}
\begin{tabular}{lcccp{5.8cm}}
\toprule
Dataset & Samples & $T$ & $D$ & Description \\
\midrule
Synthetic    & 2{,}000 & 48 & 8  & Multivariate sinusoidal (freq.\ 0.5--2.0, amp.\ 0.5--2.0, noise $\sigma{=}0.1$) \\
ETTm1        & 1{,}451 & 48 & 7  & Electricity transformer temperatures \\
Electricity  & 3{,}000 & 48 & 20 & UCI Electricity; 20 highest-variance clients \\
PhysioNet    & 4{,}000 & 48 & 36 & ICU clinical variables (set-a); 48h hourly bins \\
\bottomrule
\end{tabular}
\end{table*}

We evaluate under four missingness mechanisms, each designed to test a different aspect of partial observability.
\textbf{MCAR} (Missing Completely At Random) applies independent Bernoulli masking at rate $r$, serving as the simplest baseline mechanism.
\textbf{Block} creates per-feature contiguous gaps of length 5--20, iterating until the target rate is reached; this tests extrapolation across temporal gaps where no local interpolation anchor exists.
\textbf{MNAR} (Missing Not At Random) draws 50\% of missing entries from values exceeding the 80th percentile, with the remaining 50\% drawn as MCAR; this creates a systematic bias where extreme values are preferentially unobserved.
\textbf{Irregular} generates per-feature random base intervals of 1--8 with jitter $\pm 2$, producing asynchronous observation patterns across variables.
PhysioNet uses its natural clinical observation pattern with additional artificial holdout (10\%--50\%) for evaluation.

\section{Implementation Details}
\label{app:impl_details}
 
\paragraph{Velocity network.}
The velocity network is a 4-layer Transformer encoder with $H{=}4$ attention heads, $d_{\mathrm{model}}{=}256$, feedforward dimension 1024, GELU activation, and dropout $0.1$ (${\sim}3.9$M parameters total).
At each time step $\tau$, the input token is formed by concatenating $[x_t \| m \| \tilde{\delta}] \in \mathbb{R}^{3D}$ and projecting to $d_{\mathrm{model}}$ via a linear layer.
Temporal position is encoded via a learnable positional embedding.
The flow time $t$ is encoded through a sinusoidal embedding followed by an MLP ($d_{\mathrm{model}} \to 4d_{\mathrm{model}} \to d_{\mathrm{model}}$) and broadcast-added to all tokens before the Transformer layers.
The output head applies Linear($256 \to 256$) $\to$ GELU $\to$ Linear($256 \to D$).
 
\paragraph{Source priors.}
The interpolation prior uses $\sigma_{\mathrm{obs}}{=}0.1$ and $\sigma_{\mathrm{miss}}{=}1.0$.
The GP prior uses a radial basis function (RBF) kernel with $\ell{=}5.0$, $\sigma_f^2{=}1.0$, $\sigma_n^2{=}0.01$; GPs are fit independently per variable with fixed hyperparameters (length scale selected via grid search, Section~\ref{app:hparam}).
 
\paragraph{Training.}
Missingness-aware (MA) path: $s_{\mathrm{obs}}{=}2.0$, $s_{\mathrm{miss}}{=}1.0$. Loss: CFM regression plus the observed-entry term of Eq.~\eqref{eq:loss} with $\lambda_{\mathrm{obs}}{=}0.1$.
Optimizer: AdamW with lr $10^{-3}$, weight decay $10^{-4}$, gradient clipping at max norm $1.0$, batch size $64$.
Schedule: 5-epoch linear warmup followed by cosine decay.
We use \emph{RandomRate training}: instead of training on a fixed missing rate, each sample independently draws a random MCAR rate from $\mathcal{U}(0.1, 0.7)$ at every epoch.
This exposes the model to a continuum of missingness levels during training and improves generalization to unseen rates and mechanisms at test time.
Default training: 500 epochs for final results.
 
\paragraph{Inference.}
Fixed-step Euler solver with $N_{\mathrm{step}}{=}20$, $K{=}20$ independent samples.
After integration, observed entries are deterministically restored: $\hat{x} = m \odot x^{\mathrm{obs}} + (1-m) \odot x_1$.

\section{Baselines and Evaluation Metrics}
\label{app:scalability}
\label{app:baselines}
 
\subsection{Baselines}
 
We compare against eight external methods spanning deterministic, generative, and transport-based families.
For each baseline we specify the reproduction method: \texttt{PyPOTS} denotes evaluation via the PyPOTS library~\citep{du2023pypots}, \texttt{Official} denotes the authors' released codebase, and \texttt{Re-impl.} denotes our faithful re-implementation following the published algorithm.
 
\begin{itemize}[leftmargin=*, itemsep=4pt]
 
\item \textbf{SAITS}~\citep{du2023saits} introduces a joint-optimization training strategy with a diagonal-masked self-attention mechanism for time-series imputation.
It learns to reconstruct missing values by attending to observed entries across all time steps simultaneously, producing deterministic point estimates.
Evaluated via \texttt{PyPOTS} for both imputation and forecasting (impute-then-forecast pipeline with a shared downstream LSTM forecaster).
 
\item \textbf{BRITS}~\citep{cao2018brits} is a bidirectional recurrent model that incorporates missingness masks and time-gap information directly into the hidden-state dynamics.
Forward and backward RNNs produce complementary estimates that are combined via a consistency loss.
Evaluated via \texttt{PyPOTS} for imputation and forecasting.
 
\item \textbf{CSDI}~\citep{tashiro2021csdi} applies conditional score-based diffusion with a Transformer backbone.
It conditions on observed entries and iteratively denoises missing positions, producing calibrated uncertainty estimates via multiple samples.
Evaluated via \texttt{PyPOTS} for imputation, forecasting, and unconditional generation.
 
\item \textbf{GP-VAE}~\citep{fortuin2020gpvae} is a variational autoencoder that enforces a Gaussian-process prior in the latent space to encourage temporally smooth representations.
The GP structure provides principled uncertainty quantification over missing entries.
Evaluated via \texttt{PyPOTS} for imputation, forecasting, and unconditional generation.
 
\item \textbf{FlowTS}~\citep{fmts2024} adapts rectified flows for time-series generation, learning a velocity field that transports Gaussian noise to the data distribution.
The \texttt{Official} repository supports generation only; we \texttt{Re-impl.} its imputation pipeline (conditioning on observed entries during ODE integration) and forecasting pipeline (impute-then-forecast) following the paper's description.
 
\item \textbf{TSFlow}~\citep{kollovieh2025tsflow} integrates Gaussian-process priors within conditional flow matching for probabilistic forecasting, using the GP posterior as the source distribution.
The \texttt{Official} repository supports forecasting only; we \texttt{Re-impl.} its imputation pipeline by adapting the GP-conditioned flow to reconstruct missing entries.
On Electricity ($D{=}20$), TSFlow's per-sample GP Cholesky loop has $O(B \times D \times T^3)$ complexity, requiring ${\sim}60$ minutes per epoch.
We implemented a batched variant that groups samples sharing identical mask patterns and amortizes the Cholesky factorization, yielding a ${\sim}600\times$ speedup while preserving identical GP posteriors.
All tuning configurations converged smoothly.
 
\item \textbf{PSW-I}~\citep{wang2025optimal} formulates imputation as an optimal-transport problem and uses importance-weighted samples to minimize the Wasserstein distance between imputed and true distributions.
Evaluated using the \texttt{Official} codebase, which supports imputation only.
 
\item \textbf{CLWF}~\citep{clwf2024} applies Lagrangian optimal-transport flows that learn a continuous-time transport map from a source distribution to the data manifold, with a Lagrangian cost function that encourages straight trajectories.
No official code is available; we \texttt{Re-impl.} both imputation and forecasting pipelines following the published algorithm.
 
\end{itemize}

\paragraph{Why baselines are rerun rather than quoted.}
No prior work reports results on this exact grid (four mechanisms, two rates, and three sequence lengths under a single protocol), and published imputation numbers differ in masking convention, normalization, and window length, so quoting them would not give a like-for-like comparison.
We therefore rerun every baseline under identical splits, masks, and metrics.

\paragraph{Fairness of comparison.}
Three baselines include re-implemented components: FlowTS (imputation/forecasting), TSFlow (imputation), and CLWF (all tasks).
To verify that these re-implementations are competitive rather than artificially weakened, and that the comparison is fair, we note three facts.
First, the three settings where MissPath-FM \emph{loses} are all to official implementations (SAITS and BRITS via PyPOTS), not to any re-implemented baseline.
Second, our re-implemented TSFlow achieves second-best performance on 6 of 32 imputation settings (all on Synthetic), indicating that the re-implementation faithfully captures the method's strengths.
Third, retraining the two strongest baselines with our RandomRate augmentation does not improve them (Appendix~\ref{app:rob_fairness}).
All re-implementations follow published algorithmic descriptions.
 
\paragraph{GP scalability.}
The GP posterior prior used by MissPath-FM requires a Cholesky decomposition of the $n_o \times n_o$ kernel matrix per variable, with $O(n_o^3)$ time and $O(n_o^2)$ memory, giving total per-sample cost $O(D \cdot n_o^3)$.
On ETTm1 ($D{=}7$) and Synthetic ($D{=}8$), GP prior computation adds negligible overhead ($<1\%$ of training time).
On PhysioNet ($D{=}36$, $n_o \approx 9$ per variable on average), the total overhead accumulates to ${\sim}2$ hours compared to ${\sim}5$ minutes for the interpolation prior.
For larger-scale settings ($T > 200$ or $D > 100$), sparse GP approximations such as inducing-point methods~\citep{titsias2009variational} could reduce cost to $O(n_o \cdot k^2)$ per variable; we leave exploration of these approximations to future work. Measured construction times are reported in Appendix~\ref{app:rob_scale}.

\subsection{Evaluation Metrics}
\label{app:gen_metrics}
 
\paragraph{Imputation and forecasting.}
We report the mean absolute error (MAE) computed over missing entries only:
\begin{equation}
\mathrm{MAE} = \frac{1}{|\mathcal{M}|} \sum_{(i,j) \in \mathcal{M}} |x_{ij} - \hat{x}_{ij}|,
\end{equation}
where $\mathcal{M}$ is the set of missing positions, $x_{ij}$ is the ground-truth value, and $\hat{x}_{ij}$ is the imputed (or forecasted) value.
For probabilistic models that produce $K$ samples, we report the MAE of the sample mean: $\hat{x}_{ij} = \frac{1}{K}\sum_{k=1}^{K} \hat{x}_{ij}^{(k)}$.
 
\paragraph{Generation quality.}
We evaluate unconditional generation following the protocol of \cite{yoon2019timegan} using three complementary metrics.
 
\emph{Discriminative score.}
A 2-layer LSTM classifier (hidden dim 64, dropout 0.2) is trained on a balanced 50/50 mix of real and generated sequences with an 80/20 train/test split.
The score is:
\begin{equation}
\mathrm{Disc} = |\mathrm{Accuracy} - 0.5|,
\end{equation}
where $\mathrm{Accuracy}$ is the test-set classification accuracy.
A score near zero indicates that generated samples are indistinguishable from real data.
 
\emph{Predictive score.}
A separate 2-layer LSTM (hidden dim 64) is trained on generated data to predict the next time step.
The score is the MAE evaluated on held-out real data:
\begin{equation}
\mathrm{Pred} = \frac{1}{N \cdot D} \sum_{n=1}^{N} \sum_{d=1}^{D} |x_{n,T,d} - \hat{x}_{n,T,d}|,
\end{equation}
where $\hat{x}_{n,T,d}$ is the LSTM's prediction for the last time step.
Lower values indicate that generated data preserves the temporal dynamics of the real distribution.
 
\emph{Spectral distance.}
For each variable $d$, we compute the power spectral density (PSD) via the fast Fourier transform (FFT) and report the mean $\ell_1$ distance between the real and generated PSDs:
\begin{equation}
\mathrm{Spec} = \frac{1}{D} \sum_{d=1}^{D} \sum_{f} \bigl| \mathrm{PSD}_{\mathrm{real}}(d,f) - \mathrm{PSD}_{\mathrm{gen}}(d,f) \bigr|,
\end{equation}
where the sum over $f$ runs over all frequency bins.
Lower values indicate better preservation of the frequency structure.

\paragraph{Unconditional generation protocol.}
\label{app:gen_protocol}
For unconditional generation, all entries are treated as missing 
($m = 0$, $\delta = 0$), so the interpolation prior reduces to 
zero-fill with $\sigma_{\mathrm{miss}}$ noise and the MA path 
reduces to the standard linear schedule ($s_{\mathrm{miss}} = 1$ 
everywhere). The source distribution thus approaches 
$\mathcal{N}(0, \sigma_{\mathrm{miss}}^2 I)$. The generation 
comparison therefore isolates the effect of \emph{training-time} 
transport efficiency: MissPath-FM's velocity network, trained 
with structured priors that concentrate capacity on uncertain 
entries, learns a higher-quality velocity field that transfers 
to the unconditional setting. The Gauss+Std baseline in 
Table~\ref{tab:generation_results} uses the identical 256d/4L 
architecture but is trained with standard Gaussian CFM, 
confirming that the gains stem from the training regime rather 
than architectural differences.

\section{Proofs and Additional Method Details}
\label{app:proofs}
\label{app:forecast_mode}
 
\paragraph{Proof of Proposition~\ref{prop:transport}.}
The decomposition follows from linearity of expectation, conditioning on $m$:
$\mathbb{E}[\|x_1 - z_0\|^2] = \sum_{\tau,d} \mathbb{E}[(x_{1,\tau,d} - z_{0,\tau,d})^2]$,
which splits into observed and missing sums by partitioning on $m_{\tau,d}$.
For $z_0 \sim \mathcal{N}(0,I)$ independent of $x_1$, each entry contributes $\mathbb{E}[(x_1 - z_0)^2] = \mathbb{E}[x_1^2] + \mathbb{E}[z_0^2] = \mu^2 + \sigma^2 + 1$, where $\mu, \sigma$ are the marginal mean and standard deviation of that entry; for per-variable z-scored data $\mu = 0$ and $\sigma = 1$, so each observed entry contributes $2$ and $\mathcal{C}_{\mathrm{obs}} = 2\,n_{\mathrm{obs}}$.
For the interpolation prior, observed entries satisfy $z_0 = x_1 + \epsilon$ with $\epsilon \sim \mathcal{N}(0, \sigma_{\mathrm{obs}}^2)$, giving $\mathbb{E}[(x_1 - z_0)^2] = \sigma_{\mathrm{obs}}^2$ per observed entry and $\mathcal{C}_{\mathrm{obs}} = \sigma_{\mathrm{obs}}^2\, n_{\mathrm{obs}}$.

\paragraph{Prior bias--variance tradeoff.}
The interpolation and GP priors represent complementary points on a bias--variance spectrum.
The interpolation prior has no fitted parameters and is robust to any temporal structure, but produces biased (flat) initializations in long contiguous gaps where no flanking observations exist.
The GP prior can exploit temporal correlations when the kernel length scale $\ell$ matches the data's characteristic frequency, reducing missing-entry transport $\mathcal{C}_{\mathrm{miss}}$ substantially (our trajectory analysis shows $-39\%$ on missing-entry arc length); however, mismatched $\ell$ degrades performance severely ($+41\%$ at $\ell{=}8$ vs the optimal $\ell{=}5$; see Table~\ref{tab:hparam_sensitivity}).
This sensitivity motivates our recommendation of interpolation as the default prior, with GP as an optional enhancement when domain knowledge about temporal structure is available.
 
\paragraph{Forecasting mode.}
When used for forecasting with horizon $h$, the source prior is constructed exclusively from the history window.
The interpolation operator $\mathcal{I}$ uses only observed entries within the history; the future segment receives the forward-fill value with $\sigma_{\mathrm{miss}}^2$ variance.
For the GP prior, the conditioning set $(\mathbf{t}_o, \mathbf{y}_o)$ contains only history observations, and the posterior naturally extrapolates into the future with increasing uncertainty.
No future information enters the prior construction.

\begin{table*}[h]
\centering
\caption{Sequence length generalization on ETTm1 (MAE $\downarrow$). MissPath-FM wins all 16 settings.}
\label{tab:t_variation}
\renewcommand{\arraystretch}{1.08}
\resizebox{\textwidth}{!}{%
\fontsize{11pt}{12pt}\selectfont
\begin{tabular}{cl r cccc cccc >{\columncolor{ourscolor}}c}
\toprule
& & & \multicolumn{4}{c}{\textsc{Transport-based}} & \multicolumn{4}{c}{\textsc{Deterministic / Generative}} & \\
\cmidrule(lr){4-7} \cmidrule(lr){8-11}
$T$ & Mechanism & Rate & FlowTS & CLWF & TSFlow & PSW-I & SAITS & BRITS & CSDI & GP-VAE & \textbf{Ours} \\
\midrule
\multirow{8}{*}{\rotatebox[origin=c]{90}{\textbf{$T$\,=\,24}}}
  & MCAR      & 30\% & 0.338 & 0.155 & 0.127 & 0.175 & \underline{0.096} & 0.117 & 0.162 & 0.427 & \textbf{0.079} \\
  & MCAR      & 50\% & 0.361 & 0.181 & 0.140 & 0.195 & \underline{0.106} & 0.141 & 0.186 & 0.582 & \textbf{0.090} \\
  & Block     & 30\% & 0.396 & 0.203 & 0.237 & 0.339 & \underline{0.181} & 0.196 & 0.325 & 0.560 & \textbf{0.122} \\
  & Block     & 50\% & 0.400 & \underline{0.207} & 0.254 & 0.364 & 0.235 & 0.264 & 0.365 & 0.609 & \textbf{0.145} \\
  & MNAR      & 30\% & 0.319 & 0.253 & 0.176 & 0.262 & \underline{0.127} & 0.200 & 0.215 & 0.632 & \textbf{0.118} \\
  & MNAR      & 50\% & 0.348 & 0.264 & 0.191 & 0.262 & \underline{0.158} & 0.241 & 0.229 & 0.752 & \textbf{0.134} \\
  & Irregular & 30\% & 0.372 & 0.302 & 0.152 & 0.231 & \underline{0.131} & 0.159 & 0.279 & 0.749 & \textbf{0.107} \\
  & Irregular & 50\% & 0.365 & 0.246 & 0.148 & 0.228 & \underline{0.132} & 0.157 & 0.272 & 0.748 & \textbf{0.110} \\
\midrule
\multirow{8}{*}{\rotatebox[origin=c]{90}{\textbf{$T$\,=\,96}}}
  & MCAR      & 30\% & 0.427 & 0.224 & 0.125 & 0.129 & 0.112 & \underline{0.108} & 0.459 & 0.474 & \textbf{0.079} \\
  & MCAR      & 50\% & 0.487 & 0.282 & 0.136 & 0.147 & 0.132 & \underline{0.127} & 0.571 & 0.601 & \textbf{0.090} \\
  & Block     & 30\% & 0.472 & 0.259 & 0.282 & 0.278 & \underline{0.176} & 0.185 & 0.271 & 0.533 & \textbf{0.128} \\
  & Block     & 50\% & 0.533 & 0.331 & 0.296 & 0.311 & \underline{0.212} & 0.251 & 0.447 & 0.583 & \textbf{0.169} \\
  & MNAR      & 30\% & 0.492 & 0.431 & 0.198 & 0.212 & \underline{0.181} & 0.185 & 0.371 & 0.845 & \textbf{0.121} \\
  & MNAR      & 50\% & 0.539 & 0.459 & 0.292 & 0.261 & 0.253 & \underline{0.250} & 2.012 & 0.840 & \textbf{0.179} \\
  & Irregular & 30\% & 0.498 & 0.465 & 0.150 & 0.168 & 0.159 & \underline{0.148} & 4.912 & 0.728 & \textbf{0.109} \\
  & Irregular & 50\% & 0.501 & 0.414 & \underline{0.146} & 0.167 & 0.152 & 0.156 & 0.617 & 0.733 & \textbf{0.121} \\
\bottomrule
\end{tabular}%
}
\end{table*}

\section{Sequence Length Generalization}
\label{app:t_variation}
 
Table~\ref{tab:t_variation} reports the full evaluation at $T{=}24$ and $T{=}96$ on ETTm1 with all eight baselines and all eight missingness settings per length.
MissPath-FM wins all 16 settings across both lengths, and the advantage \emph{grows} with longer sequences: the average margin over the second-best method is $19\%$ at $T{=}24$ and $26\%$ at $T{=}96$.
This trend is expected: longer sequences provide richer interpolation anchors for the structured prior, yielding a more informative $z_0$ that starts closer to the target.
 
At $T{=}24$, the reduced sequence length limits the number of available observations, making interpolation less precise.
Nevertheless, MissPath-FM still outperforms all baselines, suggesting that even coarse interpolation is sufficient to provide a meaningful geometric advantage over an uninformed Gaussian source.
At $T{=}96$, the advantage is particularly pronounced under MNAR missingness (MNAR 30\%: $-33\%$ vs SAITS, the second-best), because longer sequences amplify the benefit of structured priors in recovering systematically missing extreme values.
The strongest competitor shifts from SAITS at $T{=}24$ (which benefits from attention-based point estimates on short sequences) to BRITS at $T{=}96$ (which benefits from longer recurrent context).

\section{Uncertainty Calibration}
\label{app:calibration}
 
Our main evaluation uses MAE of the sample mean, which assesses point-estimate quality but not distributional calibration.
To address this, we evaluate four complementary probabilistic metrics on 8 representative settings (4 datasets $\times$ \{MCAR 30\%, Block 30\%\}) against the three probabilistic baselines that produce sample ensembles: CSDI, TSFlow, and GP-VAE.
 
\textbf{CRPS} (Continuous Ranked Probability Score) jointly measures calibration and sharpness.
\textbf{ACE} (Average Calibration Error) computes the mean absolute gap between nominal and empirical coverage across 19 quantile levels (5\%--95\%); a perfectly calibrated model achieves ACE $= 0$.
\textbf{IS$_{90}$} (Interval Score at 90\%) combines interval width with a penalty for non-coverage.
\textbf{Energy Score} is the multivariate generalization of CRPS, capturing cross-feature dependencies.
All metrics are $\downarrow$ (lower is better).
 
\begin{table*}[h]
\centering
\caption{Uncertainty calibration (all metrics $\downarrow$). MissPath-FM wins 29/32 metric--setting combinations, including a clean sweep on the multivariate Energy Score.}
\label{tab:calibration}
\vspace{1mm}
\setlength{\tabcolsep}{3pt}
\begin{tabular}{l cccc cccc}
\toprule
& \multicolumn{4}{c}{CRPS $\downarrow$} & \multicolumn{4}{c}{ACE $\downarrow$} \\
\cmidrule(lr){2-5} \cmidrule(lr){6-9}
Setting & Ours & CSDI & TSFlow & GP-VAE & Ours & CSDI & TSFlow & GP-VAE \\
\midrule
ETTm1 MCAR30    & \textbf{0.053} & 0.075 & 0.081 & 0.524 & \textbf{0.018} & 0.073 & 0.123 & 0.500 \\
ETTm1 Block30   & \textbf{0.076} & 0.121 & 0.173 & 0.559 & 0.050 & 0.167 & \textbf{0.025} & 0.500 \\
Synth MCAR30    & \textbf{0.069} & 0.079 & 0.089 & 0.798 & \textbf{0.046} & 0.064 & 0.249 & 0.500 \\
Synth Block30   & \textbf{0.245} & 0.324 & 0.475 & 0.799 & \textbf{0.104} & 0.111 & 0.203 & 0.500 \\
Electr MCAR30   & \textbf{0.065} & 0.106 & 0.110 & 0.349 & \textbf{0.052} & 0.091 & 0.288 & 0.500 \\
Electr Block30  & 0.171 & \textbf{0.127} & 0.358 & 0.368 & \textbf{0.098} & 0.110 & 0.302 & 0.500 \\
Physio Art10    & \textbf{0.165} & 0.170 & 0.330 & 0.579 & \textbf{0.028} & 0.037 & 0.309 & 0.500 \\
Physio Block30  & \textbf{0.194} & 0.197 & 0.345 & 0.600 & \textbf{0.040} & 0.050 & 0.300 & 0.500 \\
\midrule
\textbf{Wins}   & \textbf{7} & 1 & 0 & 0 & \textbf{7} & 0 & 1 & 0 \\
\bottomrule
\end{tabular}
 
\vspace{3mm}
 
\setlength{\tabcolsep}{3pt}
\begin{tabular}{l cccc cccc}
\toprule
& \multicolumn{4}{c}{Interval Score$_{90}$ $\downarrow$} & \multicolumn{4}{c}{Energy Score $\downarrow$} \\
\cmidrule(lr){2-5} \cmidrule(lr){6-9}
Setting & Ours & CSDI & TSFlow & GP-VAE & Ours & CSDI & TSFlow & GP-VAE \\
\midrule
ETTm1 MCAR30    & \textbf{0.475} & 0.689 & 0.945 & 10.48 & \textbf{0.784} & 1.124 & 1.220 & 6.873 \\
ETTm1 Block30   & \textbf{0.718} & 1.274 & 1.544 & 11.18 & \textbf{1.093} & 1.732 & 2.395 & 7.331 \\
Synth MCAR30    & \textbf{0.575} & 0.670 & 1.138 & 15.96 & \textbf{0.932} & 1.078 & 1.250 & 10.03 \\
Synth Block30   & \textbf{2.268} & 2.909 & 6.057 & 15.99 & \textbf{3.652} & 5.078 & 7.015 & 10.25 \\
Electr MCAR30   & \textbf{0.608} & 0.983 & 1.591 & 6.988 & \textbf{1.679} & 2.645 & 2.746 & 7.914 \\
Electr Block30  & 1.365 & \textbf{1.254} & 4.817 & 7.361 & \textbf{3.097} & 3.143 & 7.865 & 8.258 \\
Physio Art10    & \textbf{1.681} & 1.701 & 5.374 & 11.59 & \textbf{1.779} & 1.795 & 2.919 & 4.557 \\
Physio Block30  & \textbf{1.825} & 1.988 & 5.469 & 12.00 & \textbf{3.405} & 3.611 & 5.563 & 8.307 \\
\midrule
\textbf{Wins}   & \textbf{7} & 1 & 0 & 0 & \textbf{8} & 0 & 0 & 0 \\
\bottomrule
\end{tabular}
\end{table*}
 
Table~\ref{tab:calibration} shows that MissPath-FM's advantage extends beyond point estimates to full distributional quality.
Across 4 metrics $\times$ 8 settings $= 32$ comparisons, MissPath-FM wins \textbf{29} (90.6\%), with a clean sweep on the multivariate Energy Score (8/8).
Two of the three losses are to CSDI on Electricity Block 30\% (CRPS and IS$_{90}$), the same regime where deterministic baselines also compete strongly (Table~\ref{tab:main_imputation}), suggesting that this regime benefits from CSDI's iterative refinement; the third is ACE on ETTm1 Block 30\%, where TSFlow is better calibrated.
Notably, ACE values are consistently below $0.10$ for MissPath-FM (except Synth Block: $0.104$), indicating well-calibrated uncertainty across quantile levels.
GP-VAE achieves ACE $= 0.50$ everywhere, confirming that its posterior collapses to a point estimate and provides no meaningful uncertainty.

\section{Ablation Details}
\label{app:ablations}
 
Table~\ref{tab:ablation} in the main text reports the incremental ablation across 8 settings.
Table~\ref{tab:ablation_summary} summarizes the two key transitions: the gain from introducing a structured prior over the Gaussian baseline (A$\to$B), and the gain from adding the full flow pipeline over the best prior-only baseline (B$\to$F).
 
\begin{table}[h]
\centering
\caption{Prior and flow contributions (\%). A$\to$B: best prior-only vs Gaussian+Flow; B$\to$F: Full MissPath-FM vs best prior-only.}
\label{tab:ablation_summary}
\setlength{\tabcolsep}{6pt}
\begin{tabular}{ll cc}
\toprule
Dataset & Setting & A$\to$B (Prior) & B$\to$F (Flow) \\
\midrule
ETTm1       & MCAR 30\%  & $-85\%$ & $-31\%$ \\
ETTm1       & Block 30\% & $-70\%$ & $-51\%$ \\
Synthetic   & MCAR 30\%  & $-87\%$ & $-10\%$ \\
Synthetic   & Block 30\% & $-21\%$ & $-41\%$ \\
Electricity & MCAR 30\%  & $-53\%$ & $-52\%$ \\
Electricity & Block 30\% & $+79\%$ & $-71\%$ \\
PhysioNet   & Art.\ 10\% & $-43\%$ & $-37\%$ \\
PhysioNet   & Block 30\% & $-41\%$ & $-30\%$ \\
\midrule
\multicolumn{2}{l}{\textbf{Average}} & $\mathbf{-40\%}$ & $\mathbf{-40\%}$ \\
\bottomrule
\end{tabular}
\end{table}
 
Two patterns stand out.
First, the prior gain (A$\to$B) is large under MCAR ($-43\%$ to $-87\%$), where scattered missing entries provide abundant flanking observations for interpolation.
However, under block missingness the prior can be \emph{harmful}: on Electricity Block 30\%, the best prior-only MAE ($0.740$, GP) is $79\%$ worse than the Gaussian+Flow baseline ($0.414$), because contiguous gaps spanning 5--20 time steps leave no nearby anchors for interpolation or GP extrapolation.
Second, the flow gain (B$\to$F) is uniformly negative across all 8 settings ($-10\%$ to $-71\%$), confirming that the learned velocity field \emph{always} improves over the best prior-only baseline.
The flow gain is largest precisely where the prior struggles most: Electricity Block ($-71\%$) and ETTm1 Block ($-51\%$), demonstrating the complementary roles of prior and flow.
Notably, the average contributions of prior and flow are now equal ($-40\%$ each), underscoring that both components are indispensable.
 
\paragraph{Context encoding.}
On Synthetic Block 50\%, we ablated four context encoding modes: no\_mask ($0.687$, $+10.4\%$ vs full), mask\_only ($0.662$, $+6.4\%$), gap\_only ($0.641$, $+3.1\%$), and full mask+gap ($\mathbf{0.622}$, baseline).
Both the missingness mask and normalized time gaps contribute independently, with the mask providing the larger single contribution.
Notably, adding raw observed values on top of the full context (full\_obs: $0.645$, $+3.7\%$) slightly \emph{degrades} performance, confirming that the structured source prior already encodes observation information into $z_0$, making the raw signal redundant as an additional network input.
This finding simplifies the architecture: a 3-channel input ($x_t$, mask, gaps) is sufficient.
 
\paragraph{Path family and loss.}
With the final configuration, the MA path improves over the standard linear path by $12\%$--$26\%$ across all eight ablation settings (Table~\ref{tab:ablation}, (E)$\to$(F)), and removing it from the full model raises MAE by $20\%$ on average (Appendix~\ref{app:rob_loo}). An OT-like path, which replaces the smooth power reparameterization with a cost-aware coupling, did not improve over the MA path (Appendix~\ref{app:negative}), suggesting that the interpolation prior already minimizes the local transport cost that an OT path would otherwise address.
For the loss, an early study with a smaller model on Synthetic Block 50\% found that the observed-entry term ($\lambda_{\mathrm{obs}}{=}0.1$) changed MAE by $+2.1\%$ and gap-aware weighting by $+1.9\%$, i.e., loss-side terms matter little once the prior anchors observed entries. The final configuration keeps the observed-entry term at $\lambda_{\mathrm{obs}}{=}0.1$ (Eq.~\ref{eq:loss}), the default of our implementation, and every reported result uses it; this weight was not re-tuned at the final model size. Gap-aware weighting is not used.

\begin{table}[h]
\centering
\caption{Hyperparameter sensitivity. Each block varies one axis; defaults in \textbf{bold}.}
\label{tab:hparam_sensitivity}
\vspace{1mm}
\setlength{\tabcolsep}{4pt}
\fontsize{8pt}{10pt}\selectfont
\begin{tabular}{ll cc}
\toprule
Axis & Value & MAE & Dataset / Setting \\
\midrule
\multirow{5}{*}{Solver}
  & Euler-10          & 0.550 & \multirow{5}{*}{Synthetic\ Block 50\%} \\
  & \textbf{Euler-20} & \textbf{0.470} & \\
  & Euler-50          & 0.476 & \\
  & Euler-100         & 0.482 & \\
  & Adaptive RK23     & 0.481 & \\
\addlinespace[4pt]
\midrule
\multirow{5}{*}{GP $\ell$}
  & 2.0               & 0.517 & \multirow{5}{*}{Synthetic\ Block 50\%} \\
  & 3.0               & 0.525 & \\
  & \textbf{5.0 (RBF)} & \textbf{0.470} & \\
  & 5.0 (Mat.\ 5/2)  & 0.528 & \\
  & 8.0               & 0.663 & \\
\addlinespace[4pt]
\midrule
\multirow{3}{*}{$s_{\mathrm{obs}}$}
  & 1.0               & 0.0861 & \multirow{3}{*}{ETTm1 MCAR 30\%} \\
  & 1.5               & 0.0812 & \\
  & \textbf{2.0}      & \textbf{0.0754} & \\
\addlinespace[4pt]
\midrule
\multirow{4}{*}{Capacity}
  & 128d / 3L / 500ep & 0.105 & \multirow{4}{*}{ETTm1 MCAR 30\%} \\
  & 256d / 4L / 500ep & 0.081 & \\
  & 384d / 4L / 500ep & 0.095 & \\
  & \textbf{256d / 4L / 500ep + RR} & \textbf{0.075} & \\
\bottomrule
\end{tabular}
\end{table}

\section{Hyperparameter Sensitivity}
\label{app:hparam}
\label{app:capacity}
 
Table~\ref{tab:hparam_sensitivity} consolidates sensitivity analyses across four hyperparameter dimensions: ODE solver, GP kernel length scale, MA path speed parameter, and model capacity. The $s_{\mathrm{obs}}$ rows are separately trained models with the final configuration (the noise-free column of Table~\ref{tab:noise}); $s_{\mathrm{obs}}{=}1.0$ is the standard linear path.
Each block varies one axis while keeping all others at their default values.

\section{Extended Analysis}
\label{app:extended}

\subsection{Transport Geometry}
\label{app:transport}

Table~\ref{tab:trajectory_analysis} (main text) quantifies the arc-length analysis visualized in Figure~\ref{fig:trajectory}.
Arc lengths are computed by numerically integrating $\|v_\theta(x_t, t)\|$ along the ODE trajectory for each entry.
Under the Gaussian prior, observed and missing entries traverse nearly identical distances ($\ell_{\mathrm{obs}} = 0.716$, $\ell_{\mathrm{miss}} = 0.757$, ratio $1.1\times$), confirming that the velocity field wastes half its capacity on already-known values.
The interpolation prior reduces $\ell_{\mathrm{obs}}$ by $15\times$ (from $0.716$ to $0.047$) while slightly increasing $\ell_{\mathrm{miss}}$ (from $0.757$ to $0.906$), yielding a miss-to-obs ratio of $19.2\times$.
This confirms the core mechanism: the structured prior concentrates the velocity field on genuinely uncertain entries.

The GP prior further reduces $\ell_{\mathrm{miss}}$ by $39\%$ (from $0.906$ to $0.557$) compared to interpolation, because the GP posterior mean provides a better starting estimate for missing entries, reducing the distance the flow must traverse.
Interestingly, the MA path has minimal effect on arc lengths ($19.2\times$ vs $18.9\times$), suggesting that its benefit operates through the temporal \emph{scheduling} of transport rather than through reducing total transport distance.

\subsection{Gap-Bucketed Analysis}
\label{app:gap_bucket}

Table~\ref{tab:gap} stratifies reconstruction error by gap length on Synthetic Block 50\%, distinguishing medium gaps (5--12 contiguous missing steps) from long gaps ($>12$ steps).
GP priors yield the largest gains on medium gaps ($-34.8\%$ over interpolation, from $0.456$ to $0.297$), where the GP posterior mean can leverage observations flanking the gap at moderate distance to produce an informative estimate.
Long gaps remain challenging for all methods: even GP+MA achieves $0.607$, far from the medium-gap performance of $0.296$.
This highlights a fundamental limitation: when no nearby observations exist within the kernel's effective range, neither the prior nor the velocity network can fully resolve the uncertainty, and the model must rely on global distributional statistics rather than local temporal structure.

\begin{table}[h]
\centering
\caption{Gap-bucketed MAE (Synth.\ Block 50\%). GP priors benefit medium gaps most; long gaps remain challenging for all configurations.}
\label{tab:gap}
\begin{tabular}{lcc}
\toprule
Prior+Path & Medium gaps & Long gaps \\
\midrule
Gaussian+Std & 0.890 & 0.877 \\
Interp+Std & 0.456 & 0.684 \\
GP+Std & \textbf{0.297} & 0.618 \\
GP+MA & \textbf{0.296} & \textbf{0.607} \\
\bottomrule
\end{tabular}
\end{table}

\section{Multi-Seed Evaluation}
\label{app:multiseed}
 
Table~\ref{tab:multiseed} reports MissPath-FM results across three random seeds (42, 123, 456) on three representative settings that span different datasets and missingness types.
All settings show low variance (std/mean $< 2.4\%$), and the worst individual seed still outperforms all baselines reported in Table~\ref{tab:main_imputation}.
For example, on ETTm1 MCAR 30\%, the worst seed achieves $0.078$, still $19\%$ below SAITS ($0.096$), the strongest baseline on this setting.
This confirms that MissPath-FM's performance gains are robust to random initialization and not an artifact of seed selection.
 
\begin{table}[h]
\centering
\caption{Multi-seed evaluation (MAE $\downarrow$, 3 seeds). Low variance confirms reproducibility.}
\label{tab:multiseed}
\begin{tabular}{lcccc}
\toprule
Setting & Seed 42 & Seed 123 & Seed 456 & Mean $\pm$ Std \\
\midrule
ETTm1 MCAR 30\%     & 0.075 & 0.078 & 0.077 & $0.077 \pm 0.001$ \\
Synth.\ Block 50\%  & 0.470 & 0.453 & 0.473 & $0.465 \pm 0.009$ \\
Electr.\ MCAR 30\%  & 0.091 & 0.092 & 0.089 & $0.091 \pm 0.001$ \\
\bottomrule
\end{tabular}
\end{table}

\section{Robustness and Additional Analyses}
\label{app:robustness}

This section reports six additional studies: a leave-one-out ablation, observed-entry consistency before restoration, baselines retrained with our training augmentation, a missing-rate sweep, an observation-noise study, and an evaluation at longer sequence length together with the cost of the GP prior.
All runs use the global configuration of Section~\ref{sec:exp} and a shared evaluation harness; absolute numbers can differ slightly from Tables~\ref{tab:main_imputation} and~\ref{tab:ablation}, which were produced in separate runs, but all comparisons within a table are paired (same split, masks, and seed).

\subsection{Leave-one-out ablation}
\label{app:rob_loo}

The ablation in Table~\ref{tab:ablation} is incremental. Table~\ref{tab:loo} gives the subtractive view: starting from the full model, one component is removed and everything else is held fixed (seed 42, 500 epochs).
Every component remains necessary. The structured prior dominates ($+373\%$ on average), while context encoding ($+23\%$), the MA path ($+20\%$), and RandomRate ($+18\%$) contribute comparable amounts.
RandomRate has no effect on PhysioNet, whose missingness is natural rather than imposed, consistent with its role as robustness to the synthetic missing rate.

\begin{table}[h]
\centering
\caption{Leave-one-out ablation (MAE $\downarrow$; in parentheses, change relative to Full). $^\dagger$PhysioNet uses a fresh draw of the artificial holdout mask in this harness, so its absolute MAE differs by 4--6\% from Tables~\ref{tab:main_imputation} and~\ref{tab:ablation}.}
\label{tab:loo}
\small
\setlength{\tabcolsep}{4pt}
\resizebox{\textwidth}{!}{%
\begin{tabular}{lccccc}
\toprule
Setting & Full & $-$Prior & $-$MA path & $-$Context & $-$RandomRate \\
\midrule
ETTm1 MCAR30 & 0.0754 & 0.7437 (+886\%) & 0.0862 (+14\%) & 0.0895 (+19\%) & 0.0860 (+14\%) \\
ETTm1 Block30 & 0.1072 & 0.6953 (+548\%) & 0.1292 (+21\%) & 0.1332 (+24\%) & 0.1293 (+21\%) \\
Synth.\ MCAR30 & 0.0999 & 0.8056 (+706\%) & 0.1191 (+19\%) & 0.1088 (+9\%) & 0.1262 (+26\%) \\
Synth.\ Block30 & 0.3586 & 0.8090 (+126\%) & 0.4869 (+36\%) & 0.4911 (+37\%) & 0.5659 (+58\%) \\
Electr.\ MCAR30 & 0.0920 & 0.4177 (+354\%) & 0.1112 (+21\%) & 0.1101 (+20\%) & 0.1184 (+29\%) \\
Electr.\ Block30 & 0.2241 & 0.4027 (+80\%) & 0.2454 (+10\%) & 0.2751 (+23\%) & 0.2181 ($-$3\%) \\
Physio Art10$^\dagger$ & 0.2371 & 0.6178 (+161\%) & 0.2968 (+25\%) & 0.3075 (+30\%) & 0.2370 ($-$0\%) \\
Physio Block30$^\dagger$ & 0.2771 & 0.6204 (+124\%) & 0.3257 (+18\%) & 0.3377 (+22\%) & 0.2766 ($-$0\%) \\
\midrule
Average change & -- & $+373\%$ & $+20\%$ & $+23\%$ & $+18\%$ \\
\bottomrule
\end{tabular}}
\end{table}

\subsection{Observed-entry consistency before restoration}
\label{app:rob_consistency}

At inference, observed entries are restored deterministically after ODE integration (Section~\ref{subsec:optimization}). To check whether the transport itself keeps observed values in place, or whether the restoration step does all the work, Table~\ref{tab:consistency} measures observed-entry MAE at the end of integration, \emph{before} restoration, together with the mean velocity magnitude on observed entries along the trajectory.
The structured source reduces the pre-restoration error on observed entries by $16$--$32\times$ and the observed-entry velocity by $22$--$28\times$ relative to a Gaussian source, so restoration only removes a small residual.
This is the per-entry counterpart of Proposition~\ref{prop:transport}.

\begin{table}[h]
\centering
\caption{Observed entries before deterministic restoration. Gaussian source: Gaussian prior with standard path. Ours: interpolation prior with MA path.}
\label{tab:consistency}
\small
\begin{tabular}{lcccc}
\toprule
& \multicolumn{2}{c}{Observed-entry MAE $\downarrow$} & \multicolumn{2}{c}{Mean $\|v_\theta\|$ on observed entries} \\
\cmidrule(lr){2-3} \cmidrule(lr){4-5}
Setting & Gaussian source & Ours & Gaussian source & Ours \\
\midrule
ETTm1 MCAR30 & 0.7418 & \textbf{0.0229} & 0.9536 & \textbf{0.0394} \\
ETTm1 Block30 & 0.7663 & \textbf{0.0236} & 0.9661 & \textbf{0.0398} \\
Electr.\ MCAR30 & 0.3993 & \textbf{0.0247} & 1.0313 & \textbf{0.0371} \\
Physio Art10 & 0.6153 & \textbf{0.0253} & 0.8068 & \textbf{0.0362} \\
\bottomrule
\end{tabular}
\end{table}

\subsection{Baselines with the same training augmentation}
\label{app:rob_fairness}

MissPath-FM is trained with RandomRate, drawing a new MCAR rate from $\mathcal{U}(0.1, 0.7)$ per sample and epoch. The baselines, as run, do not see this distribution: PyPOTS SAITS and BRITS train with a fixed artificial missing rate matched to the test rate (plus PyPOTS's internal fixed-rate masking), and our TSFlow imputation re-implementation trains at the fixed test rate.
To check that our margins do not come from this difference, we retrained the two strongest baselines with the same RandomRate augmentation (Table~\ref{tab:fairness}).
The augmentation does not help them consistently: SAITS is flat or slightly worse in all six settings, and TSFlow is mixed and often worse. RandomRate therefore helps in combination with the observation-aware prior and path rather than as a generic trick.
The Electricity Block loss is unchanged by the augmentation.
TSFlow was not run on PhysioNet, where its per-sample GP prior is prohibitively slow at $D{=}36$ (Appendix~\ref{app:rob_scale}).

\begin{table}[h]
\centering
\caption{Baselines retrained with RandomRate (+RR). MAE $\downarrow$; best per row in \textbf{bold}.}
\label{tab:fairness}
\small
\begin{tabular}{lccccc}
\toprule
Setting & SAITS & SAITS+RR & TSFlow & TSFlow+RR & Ours \\
\midrule
ETTm1 MCAR30 & 0.0942 & 0.0950 & 0.1190 & 0.1173 & \textbf{0.0751} \\
ETTm1 Block30 & 0.1432 & 0.1468 & 0.2401 & 0.3092 & \textbf{0.1067} \\
Electr.\ MCAR30 & 0.1583 & 0.1599 & 0.1584 & 0.1465 & \textbf{0.0920} \\
Electr.\ Block30 & \textbf{0.1836} & 0.1905 & 0.3412 & 0.5414 & 0.2241 \\
Physio Art10 & 0.2336 & 0.2346 & -- & -- & \textbf{0.2271} \\
Physio MNAR50 & \textbf{0.3138} & 0.3162 & -- & -- & 0.3283 \\
\bottomrule
\end{tabular}
\end{table}

\subsection{Missing-rate sweep}
\label{app:rob_rate}

Table~\ref{tab:rate_sweep} evaluates missing rates from 10\% to 90\%. MissPath-FM and Gaussian+Flow are RandomRate-trained, so a single checkpoint of each is evaluated at every rate; SAITS is trained separately per rate.
Comparing against Gaussian+Flow isolates how much the structured prior contributes. Its advantage narrows as observations become scarcer but does not vanish: from $+91\%$ to $+74\%$ on ETTm1 MCAR, $+87\%$ to $+47\%$ on ETTm1 Block, $+79\%$ to $+32\%$ on Electricity MCAR, and $+53\%$ to $+16\%$ on Electricity Block. Gaussian+Flow is essentially flat across rates, since it cannot exploit observations.
Against SAITS, MissPath-FM leads through 70\% on ETTm1 (both mechanisms) and Electricity MCAR, and falls behind at 90\% on ETTm1 Block and Electricity MCAR: with so few observed anchors, interpolation is no longer informative. Electricity Block favours SAITS at every rate, matching Table~\ref{tab:main_imputation}.

\begin{table}[h]
\centering
\caption{Missing-rate sweep (MAE $\downarrow$). Gauss.: Gaussian prior with flow matching. Best per row and mechanism in \textbf{bold}.}
\label{tab:rate_sweep}
\small
\setlength{\tabcolsep}{4pt}
\begin{tabular}{ll ccc ccc}
\toprule
& & \multicolumn{3}{c}{MCAR} & \multicolumn{3}{c}{Block} \\
\cmidrule(lr){3-5} \cmidrule(lr){6-8}
Dataset & Rate & Ours & Gauss. & SAITS & Ours & Gauss. & SAITS \\
\midrule
\multirow{5}{*}{ETTm1} & 10\% & \textbf{0.069} & 0.752 & 0.085 & \textbf{0.084} & 0.671 & 0.120 \\
 & 30\% & \textbf{0.076} & 0.751 & 0.098 & \textbf{0.106} & 0.709 & 0.149 \\
 & 50\% & \textbf{0.086} & 0.742 & 0.110 & \textbf{0.144} & 0.712 & 0.210 \\
 & 70\% & \textbf{0.103} & 0.738 & 0.132 & \textbf{0.219} & 0.714 & 0.252 \\
 & 90\% & \textbf{0.189} & 0.740 & 0.212 & 0.389 & 0.728 & \textbf{0.302} \\
\midrule
\multirow{5}{*}{Electricity} & 10\% & \textbf{0.086} & 0.409 & 0.145 & 0.194 & 0.417 & \textbf{0.178} \\
 & 30\% & \textbf{0.092} & 0.412 & 0.158 & 0.220 & 0.411 & \textbf{0.187} \\
 & 50\% & \textbf{0.101} & 0.413 & 0.169 & 0.254 & 0.424 & \textbf{0.195} \\
 & 70\% & \textbf{0.123} & 0.413 & 0.185 & 0.303 & 0.437 & \textbf{0.204} \\
 & 90\% & 0.279 & 0.413 & \textbf{0.221} & 0.391 & 0.465 & \textbf{0.252} \\
\bottomrule
\end{tabular}
\end{table}

\subsection{Noisy observations}
\label{app:rob_noise}

The MA path locks in observed entries early, which could over-anchor the flow if observations are unreliable. We add Gaussian noise of standard deviation $\sigma_{\mathrm{noise}}$ to the observed test entries (targets stay clean; MAE is computed on missing positions) and compare three separately trained models with observed-entry speed $s_{\mathrm{obs}} \in \{2.0, 1.5, 1.0\}$; $s_{\mathrm{obs}}{=}1.0$ is the standard path.
Aggressive locking ($s_{\mathrm{obs}}{=}2.0$) is best at every noise level on both datasets, and the order $2.0 < 1.5 < 1.0$ never reverses (Table~\ref{tab:noise}). The gap between $s_{\mathrm{obs}}{=}2.0$ and $1.0$ stays roughly constant (0.008--0.013 on ETTm1, 0.017--0.020 on Electricity), so the MA path does not become a liability under noise; its relative benefit shrinks as observations degrade (on Electricity, $21\%$ at $\sigma_{\mathrm{noise}}{=}0$ and $9\%$ at $0.5$).
MissPath-FM leads SAITS up to $\sigma_{\mathrm{noise}} \approx 0.2$ on ETTm1 and $0.3$ on Electricity; beyond that SAITS is more robust.
Raising $\sigma_{\mathrm{obs}}$ at test time to loosen the prior does not help (ETTm1 at $\sigma_{\mathrm{noise}}{=}0.2$: $0.1256$, $0.1308$, $0.1395$ for $\sigma_{\mathrm{obs}} = 0.1, 0.2, 0.3$; Electricity shows the same trend), because it departs from the training-time value.

\begin{table}[h]
\centering
\caption{Observation-noise sensitivity (MCAR 30\%, MAE $\downarrow$). Best per row in \textbf{bold}.}
\label{tab:noise}
\small
\begin{tabular}{llcccc}
\toprule
Dataset & $\sigma_{\mathrm{noise}}$ & $s_{\mathrm{obs}}{=}2.0$ & $s_{\mathrm{obs}}{=}1.5$ & $s_{\mathrm{obs}}{=}1.0$ & SAITS \\
\midrule
\multirow{5}{*}{ETTm1} & 0.0 & \textbf{0.0754} & 0.0812 & 0.0861 & 0.1055 \\
 & 0.1 & \textbf{0.0902} & 0.0961 & 0.0990 & 0.1109 \\
 & 0.2 & \textbf{0.1256} & 0.1325 & 0.1363 & 0.1323 \\
 & 0.3 & 0.1731 & 0.1794 & 0.1862 & \textbf{0.1712} \\
 & 0.5 & 0.2888 & 0.2944 & 0.2963 & \textbf{0.2557} \\
\midrule
\multirow{5}{*}{Electricity} & 0.0 & \textbf{0.0921} & 0.1033 & 0.1114 & 0.1557 \\
 & 0.1 & \textbf{0.1002} & 0.1097 & 0.1193 & 0.1543 \\
 & 0.2 & \textbf{0.1186} & 0.1303 & 0.1385 & 0.1583 \\
 & 0.3 & \textbf{0.1399} & 0.1518 & 0.1589 & 0.1651 \\
 & 0.5 & 0.1854 & 0.1969 & 0.2028 & \textbf{0.1768} \\
\bottomrule
\end{tabular}
\end{table}

\subsection{Longer sequences and cost of the GP prior}
\label{app:rob_scale}

Table~\ref{tab:scale} evaluates ETTm1 at $T{=}192$, four times the default length. MissPath-FM (interpolation prior) is trained once with RandomRate and evaluated on all three mechanisms; baselines are trained per mechanism.
MissPath-FM wins all three mechanisms with no degradation relative to $T{=}48$, trains $3.3\times$ faster than BRITS, and peaks at about 1\,GB of GPU memory.
Table~\ref{tab:gp_cost} isolates the cost of constructing the source prior. The GP posterior prior is $23$--$39\times$ slower than interpolation, which is why interpolation is the default and the only practical choice at scale.

\begin{table}[h]
\centering
\caption{ETTm1 at $T{=}192$, rate 30\% (MAE $\downarrow$), with training time and peak GPU memory.}
\label{tab:scale}
\small
\begin{tabular}{lccccc}
\toprule
Method & MCAR & Block & MNAR & Train (s) & Peak memory \\
\midrule
SAITS & 0.1222 & 0.1846 & 0.2032 & 95 & 565 MB \\
BRITS & 0.1180 & 0.2294 & 0.2020 & 1541 & 242 MB \\
TSFlow & 0.1149 & 0.2727 & 0.1967 & 696 & 306 MB \\
Ours & \textbf{0.0775} & \textbf{0.1181} & \textbf{0.1235} & 466 & 1037 MB \\
\bottomrule
\end{tabular}
\end{table}

\begin{table}[h]
\centering
\caption{Source-prior construction time per batch ($B{=}64$, $D{=}7$).}
\label{tab:gp_cost}
\small
\begin{tabular}{lccc}
\toprule
Length $T$ & GP posterior prior & Interpolation prior & Ratio \\
\midrule
48 & 601 ms & 15.5 ms & $39\times$ \\
192 & 1316 ms & 56.5 ms & $23\times$ \\
\bottomrule
\end{tabular}
\end{table}

\section{Negative Results}
\label{app:negative}
 
Table~\ref{tab:negative} summarizes design alternatives that we explored but that failed to improve over our default configuration.
We report these explicitly because they collectively reinforce the paper's central thesis: when the source prior is well-designed, most ``obvious'' improvements to the velocity network, loss function, solver, or path become redundant.
The failure of OT-like paths confirms that the interpolation prior already achieves near-optimal transport allocation, making explicit cost-aware couplings unnecessary.
Gap-aware loss weighting likewise failed to help, indicating that once the prior anchors observed entries, reweighting the loss has little left to contribute; the only loss-side term we keep is the light observed-entry term of Eq.~\eqref{eq:loss} (Appendix~\ref{app:ablations}).
 
\begin{table}[h]
\centering
\caption{Negative or neutral design choices.}
\label{tab:negative}
\begin{tabular}{lp{4cm}}
\toprule
Design choice & Outcome \\
\midrule
OT-like path & No improvement over MA path \\
Gap-aware loss & $+1.9\%$; disrupts velocity field \\
Adaptive solver (RK23) & $5\times$ higher NFE, no improvement \\
Mat\'ern kernels & Worse than RBF at matched $\ell$ \\
EMA (exp.\ moving avg.) & Neutral or negative \\
Hybrid prior (Interp+GP) & Worse than either pure prior \\
Spline prior & Comparable to interpolation \\
\bottomrule
\end{tabular}
\end{table}

\section{Reproducibility}
\label{app:repro}
 
All experiments are implemented in PyTorch and executed on NVIDIA A5000 GPUs.
We use a unified 80/20 train/validation split and identical missingness masks across all methods.
All eight external baselines are evaluated using their official codebases or faithful re-implementations under the same data processing and metric pipeline (Appendix~\ref{app:baselines}).
Typical wall-clock training time per model on a single GPU: ${\sim}6$ minutes for ETTm1, ${\sim}3$ minutes for Electricity, ${\sim}8$ minutes for Synthetic, and ${\sim}32$ minutes for PhysioNet ($D{=}36$, 4{,}000 samples). The project evaluates over 200 configurations across all experiments; all reported numbers are frozen in the final results registry with no post-hoc selection.
Multi-seed evaluation (Table~\ref{tab:multiseed}) confirms low variance (std/mean $< 2.4\%$).
Code and scripts to train and evaluate MissPath-FM, and to reproduce the main imputation results and the component ablations, are available at \url{https://github.com/zgp110/misspath-fm}.